\documentclass{article}

 \usepackage[preprint]{neurips_2026}

\usepackage[utf8]{inputenc} %
\usepackage[T1]{fontenc}    %
\usepackage{hyperref}       %
\usepackage{url}            %
\usepackage{booktabs}       %
\usepackage{amsfonts}       %
\usepackage{nicefrac}       %
\usepackage{microtype}      %
\usepackage[table]{xcolor}         %
\usepackage{graphicx}
\usepackage{subcaption}
\usepackage{algorithm}
\usepackage{algpseudocode}
\usepackage{bm}
\usepackage{placeins}
\usepackage{enumitem}
\usepackage{wrapfig}
\usepackage{amsmath}
\usepackage{tcolorbox}

\newcommand{\beps}{\bm{\epsilon}}
\newcommand{\btheta}{\bm{\theta}}

\renewcommand{\paragraph}[1]{\noindent\textbf{#1}\hspace{0.2cm}}

\newcommand{\hl}[1]{\underline{\textbf{#1}}}

\definecolor{theblue}{RGB}{30, 30, 180}
\definecolor{mygreen}{RGB}{202,228,192}
\definecolor{myblue}{RGB}{184,207,217}
\definecolor{myorange}{RGB}{237,141,91}

\hypersetup{
   colorlinks=true,
   linkcolor=[RGB]{30, 30, 180},
   citecolor=[RGB]{30, 30, 180},
   urlcolor=magenta,
   pdfborder=0 0 0,
   pdftitle={},
   pdfsubject={}, 
   pdfkeywords={},
   pdfauthor={},%
   pdfstartview=FitH
}

\title{Overcoming Forgetting in LLM Fine-Tuning with Evolution Strategies}

\author{%
  Kajetan Schweighofer\\
  Cognizant AI Lab \\
  \texttt{kai.schweighofer@gmx.at} \\
  \AND
  Conor F. Hayes\\
  Cognizant AI Lab \\
  \texttt{conor.hayes@cognizant.com} \\
  \And
  Roberto Dailey\\
  Cognizant AI Lab \\
  \texttt{roberto.dailey@cognizant.com} \\
  \And
  Risto Miikkulainen \\
  UT Austin \& Cognizant AI Lab \\
  \texttt{risto@cs.utexas.edu} \\
  \And
  Xin Qiu\\
  Cognizant AI Lab \\
  \texttt{qiuxin.nju@gmail.com} \\
}

\begin{document}

\maketitle

\begin{abstract}
  Evolution Strategies (ES) has recently emerged as a competitive alternative to reinforcement learning (RL) for large language model (LLM) fine-tuning, offering advantages through simplicity, scalability, and inference-only training. 
  However, recent work suggests that ES fine-tuning on new tasks may induce forgetting of prior tasks.
  First, this paper shows that prior task forgetting (1) is better characterized as performance drift rather than irreversible forgetting, with prior-task performance often recovering during ES training; and (2) is not a specific failure mode of ES, but can also arise for fine-tuning with RL methods. 
  Second, it analyzes when and why such drift arises, highlighting its dependence on ES training dynamics, particularly random walk behavior in weakly constrained directions of the weight space. 
  Third, based on these insights, it introduces \emph{Anchored Weight Decay} (AWD) as a parameter-space regularization technique that constrains optimization toward the initial model parameters.
  AWD effectively stabilizes prior-task performance while preserving target-task performance, achieving benefits comparable to large ES population sizes at much lower computational cost.
  Thus, contrary to previous beliefs, the paper shows that prior-task forgetting under ES is largely avoidable, positioning ES as a promising approach for continual learning in LLMs.
\end{abstract}

\renewcommand{\thefootnote}{}
\footnotetext{The code to reproduce all experiments is available at \url{https://github.com/kschweig/es-awd}\vspace{-0.4cm}}
\renewcommand{\thefootnote}{\arabic{footnote}} 

\section{Introduction}

Ever-increasing amounts of compute are spent on post-training of LLMs, adapting them to particular tasks through reward signals \citep{Cobbe:21, Guo:25}.
Reinforcement Learning (RL), in particular Group Relative Policy Optimization (GRPO) \citep{Shao:24}, has become the standard approach in this setting. 
However, RL-based training is complex with high engineering effort and limited scalability.
Recently, Evolution Strategies (ES) have reemerged as a promising alternative for LLM post-training \citep{Salimans:17, Qiu:25}. 
ES optimizes model parameters through stochastic perturbations in weight space, requiring only forward passes.
This makes ES particularly attractive for large-scale deployment, as it can leverage optimized inference systems and parallelize naturally across population members.

While ES has been shown to be competitive with, and in some cases outperform, RL-based fine-tuning \citep{Qiu:25, Hoy:26, Gan:26}, recent work reports that ES induces forgetting of previously learned tasks \citep{Abdi:26, Hoy:26}. 
These results raise concerns about the suitability of ES for continual learning scenarios, where preserving prior capabilities is essential.
However, despite demonstrating this effect, the nature, causes, and generality of the phenomenon remain insufficiently understood.
Key open questions include whether it reflects irreversible loss of knowledge or a transient effect, whether similar forgetting can arise in RL methods, how it depends on the target task and model family, and whether the larger magnitude of weight updates in ES are indeed the primary cause (as suggested by \citet{Abdi:26}).
Moreover, given that forgetting arises in at least some settings, an important next step is to develop effective mitigation strategies.

\begin{figure}
    \centering
    \begin{subfigure}{0.39\linewidth}
        \includegraphics[width=\linewidth, trim=0 0.2cm 0 0.2cm, clip]{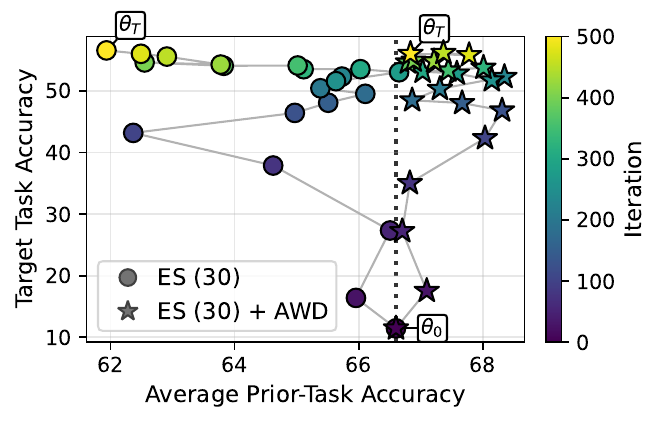}
        \subcaption{AWD mitigates forgetting}
    \end{subfigure}
    \begin{subfigure}{0.28\linewidth}
        \includegraphics[width=\linewidth, trim=0 -1.2cm 0 0, clip]{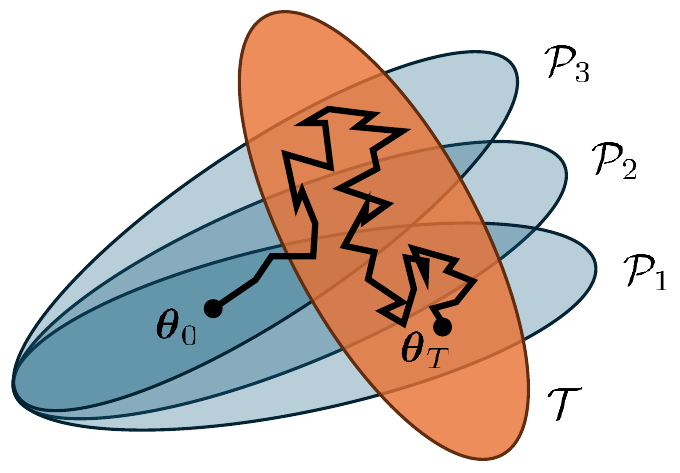}
        \subcaption{Standard ES optimization}
    \end{subfigure}
    \begin{subfigure}{0.28\linewidth}
        \includegraphics[width=\linewidth, trim=0 -1.2cm 0 0, clip]{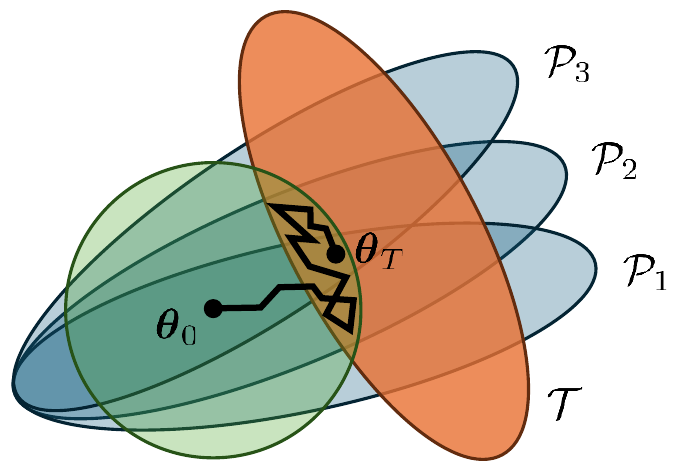}
        \subcaption{ES with AWD}
    \end{subfigure}
    \caption{\emph{Anchored Weight Decay (AWD) mitigates prior task forgetting.} (a) Target task accuracy (Countdown) vs. average prior task accuracy. The dotted line denotes prior task accuracy for the original model with weights $\btheta_0$. Standard ES shows unstable prior-task accuracy during training iterations (denoted by color), while AWD stabilizes it. (b) Prior tasks $\mathcal{P}_i$ and target task $\mathcal{T}$ occupy different high-performance regions in weight space. ES exploration within $\mathcal{T}$ can cause drift on prior tasks. (c) AWD constrains optimization near the initial model, mitigating forgetting.}
    \label{fig:main}
\end{figure}

This paper provides a comprehensive analysis of prior-task forgetting under ES fine-tuning over a diverse set of target tasks, prior tasks, model families and model sizes. 
It shows that forgetting is better characterized as \emph{performance drift} rather than irreversible forgetting, as prior-task performance often recovers during training.
The experiments further show that forgetting also occurs with RL methods, indicating that forgetting is a broader issue in LLM fine-tuning rather than one specific to ES.
For ES, its training dynamics are experimentally identified as the main driver of forgetting, in particular, a random walk in directions of weight space that are weakly constrained by the target task.
Building on this insight, the paper introduces \emph{Anchored Weight Decay (AWD)}, a simple modification to the ES update that constrains optimization toward the initial model parameters. 
As illustrated in Fig.~\ref{fig:main}, AWD reduces task-irrelevant drift while retaining the benefits of ES, stabilizing prior-task performance without sacrificing target-task accuracy.

\section{Background and Related Work}

Two methods for fine-tuning LLMs are considered in this work: ES, a population-based, gradient-free approach that samples in weight space, and GRPO, a policy-gradient RL method that samples in token space.
An introduction to ES is given in Sec.~\ref{sec:es}, to GRPO in App.~\ref{app:grpo}.
Moreover, a discussion of continual learning and prior-task forgetting, particularly in LLMs, is provided in Sec.~\ref{sec:cl}.

\subsection{Evolution Strategies} \label{sec:es}

Evolution Strategies (ES) are a class of population-based, zeroth-order optimization methods that update model parameters via stochastic perturbations and reward aggregation, rather than gradient-based backpropagation.
The objective of ES is to choose weights $\btheta$ that maximize the expected reward under Gaussian perturbations: $\max_{\btheta} \mathbb{E}_{\beps \sim \mathcal{N}(\bm{0}, I)} \left[ R(\btheta + \sigma \beps) \right]$.
Given model parameters $\btheta$ and a reward function $R(\cdot)$, ES samples perturbations $\beps_n \sim \mathcal{N}(\bm{0}, I)$, scales them by $\sigma$, and evaluates reward of perturbed models $r_n = R(\btheta + \sigma \beps_n)$.
Subsequently, rewards are normalized, for example via z-scaling over the population $\hat{r}_n = \tfrac{r_n - \mu_N}{\sigma_N}$, where $\mu_N$ and $\sigma_N$ are the mean and standard deviation over the population \citep{Qiu:25}.
The weight update then aggregates these perturbations weighted by their normalized rewards $\hat{r}_n$, through the following update rule:
\begin{equation} \label{eq:es}
    \btheta_{t} = \btheta_{t - 1} + \alpha \cdot \frac{1}{N} \sum_{n=1}^{N} \hat{r}_n \beps_n \ ,
\end{equation}
where $\alpha$ is the stepsize of the update.
The full algorithm is provided in Algorithm~\ref{alg:basic_es}.
Eq.~\eqref{eq:es} can be interpreted as a Monte Carlo estimate of the gradient of an objective that is Gaussian-smoothed in weight space \citep{Qiu:25}.
Classical ES methods date back to \citet{Rechenberg:73} and \citet{Schwefel:77}, with modern variants such as NES \citep{Wierstra:08, Wierstra:14} and CMA-ES \citep{Hansen:01} improving search efficiency.
Further, \citet{Salimans:17} demonstrated that ES can scale to deep neural networks and achieve competitive performance with RL in control tasks.

Recently, ES has been revisited as an alternative paradigm for fine-tuning LLMs \citep{Qiu:25, Korotyshova:25, Sarkar:25}, particularly in settings where only outcome-level rewards are available.
Unlike RL methods that explore in token space, ES performs exploration in weight space, allowing it to naturally handle long-horizon and sparse reward signals while avoiding token-level credit assignment.
Surprisingly, fine-tuning billion-parameter models is possible with very small population sizes; this phenomenon has been linked to low-dimensional curvature of fine-tuning landscapes \citep{Liang:26}.
Moreover, ES relies solely on forward passes thus can utilize specialized hardware and software for inference and is easily parallelized.

\subsection{Continual learning and prior-task forgetting} \label{sec:cl}

Catastrophic forgetting was first identified as an issue by \citet{McCloskey:89}, who showed that sequential training leads to rapid loss of previously learned knowledge in neural networks.
To address this, replay-based methods \citep{Lopez-Paz:17, Autume:19} store and reuse past examples during training on new tasks.
In contrast, regularization-based approaches \citep{Kirkpatrick:17, Zenke:17} constrain changes to model parameters important for prior tasks.
These ideas are closely related to the proposed method AWD as discussed in App.~\ref{app:ewc}.
Another line of work freezes previously learned parameters and expands the model architecture with additional parameters that are trained on new tasks \citep{Rusu:16}.
A comprehensive survey of continual learning and prior-task forgetting is given by \citet{Parisi:19}.

Recent work has examined prior-task forgetting in the context of LLMs.
\citet{Scialom:22} showed that sequential instruction tuning leads to performance degradation across tasks such as summarization and style transfer, highlighting instability in cross-task generalization.
Complementing this, \citet{Luo:25} provided a large-scale empirical study demonstrating that catastrophic forgetting consistently occurs during continual fine-tuning of LLMs, affecting domain knowledge, reasoning, and comprehension, and even increasing with model scale.
To mitigate these effects, \citet{Wang:23} proposed O-LoRA, which learns task-specific updates in orthogonal low-rank subspaces to reduce interference between tasks.
\citet{Wu:24} adopted an architectural strategy that expands transformer blocks for new tasks while freezing existing parameters, increasing model capacity to preserve prior knowledge.
\citet{Shenfeld:25, Chen:25, Lai:26} investigated forgetting of prior tasks under RL and supervised fine-tuning, finding that RL methods lead to less forgetting compared to supervised fine-tuning.

Most closely related to this paper are the recent studies by \citet{Abdi:26} and \citet{Hoy:26} on prior-task forgetting under ES fine-tuning.
\citet{Abdi:26} demonstrate that ES fine-tuning on a target task can lead to prior-task forgetting, attributing it to larger weight updates compared to GRPO.
This paper extends their setting by systematically investigating different target tasks, prior tasks and model architectures, as well as the influence of ES population size, thus clarifying the reason for larger magnitudes of weight updates, and ultimately introducing AWD as mitigation strategy for forgetting.
\citet{Hoy:26} investigated forgetting of ES vs. GRPO in a sequential learning setting on multiple tasks.
They provided an analysis of the geometry of weight changes, showing that GRPO and ES solutions are linearly mode connected though their update directions per iteration are near-orthogonal.
Furthermore, they suggest a theoretical model of optimization with ES, showing that ES induces a random walk in directions of the weight space that are weakly coupled to the target task. 
To mitigate forgetting, they suggest early stopping, balancing prior task forgetting with performance on the target task.
This paper further investigates the relationship between the random walk behavior of ES and prior-task forgetting, motivating AWD as a simple yet effective method to mitigate forgetting.

\section{Experimental Setup} \label{sec:setup}

All experiments in this paper follow a common setup designed to study prior-task forgetting. 
This setup includes multiple complementary prior tasks and three diverse target tasks for training.

\paragraph{Target tasks.}
Three different target tasks are considered, Countdown \citep{Gandhi:24, Pan:25}, GSM8K \citep{Cobbe:21} and ProofWriter \citep{Tafjord:21}.
These tasks focus on arithmetic reasoning, multi-step numerical reasoning, and logical reasoning, respectively.
Training was performed for a fixed set of 200 examples.

\paragraph{Prior tasks.}
As prior tasks, HellaSwag \citep{Zellers:19}, PIQA \citep{Bisk:20}, ARC-Challenge \citep{Clark:18} and MMLU-Pro \citep{Wang:24} are considered.
They focus on selecting plausible sentence continuations, commonsense reasoning, and professional and academic knowledge across multiple disciplines.
Furthermore, target tasks not trained on within an experiment are considered as additional prior tasks.

\paragraph{Evaluation.}
Evaluation was conducted on 500–2,000 examples per task, depending on how many were available; for target tasks, evaluation sets were disjoint from the training data. 
When more than 2,000 examples were available, a random subset of 2,000 was sampled.
Greedy decoding is used for evaluation and accuracy reported. 

\paragraph{Model.}
In line with \citet{Abdi:26}, who first identified forgetting through ES fine-tuning, the Qwen-2.5 3B Instruct model \citep{Qwen2.5} is used, unless stated otherwise. 
The chat template is applied, and the maximum generated sequence length capped at 1024 tokens, which is sufficient to generate valid reasoning and answers for all tasks considered.

\paragraph{Implementations of ES and GRPO.}
The ES implementation follows \citet{Qiu:25}, using their optimized variant leveraging vLLM \citep{Kwon:23} for efficient sequence generation and scoring of individual members.
For GRPO, the implementation provided through the \texttt{verl} framework \citep{Sheng:24} was used. 
This setup is consistent with \citet{Abdi:26}, who used the same implementations.
ES is used with population size 30 if not otherwise specified, GRPO is used with KL penalty. 
Additional details on the experimental setup are provided in App.~\ref{app:evaluation}.

\section{Understanding Forgetting in LLM Fine-Tuning} \label{sec:analysis}

To motivate a mitigation strategy for prior-task forgetting in LLM fine-tuning with ES, it is first examined when and why such forgetting arises.
Specifically, the forgetting behavior of ES and GRPO is compared throughout the training process.
Furthermore, the dependence of forgetting on the target task, prior task and model types is examined.
Finally, the role of ES training dynamics in driving forgetting is investigated, suggesting a mitigation strategy.

\subsection{Forgetting throughout the training process} \label{sec:forgetting_during_training}

\begin{figure}[b]
    \centering
    \includegraphics[width=\linewidth, trim=0.2cm 0.2cm 0.2cm 0.2cm, clip]{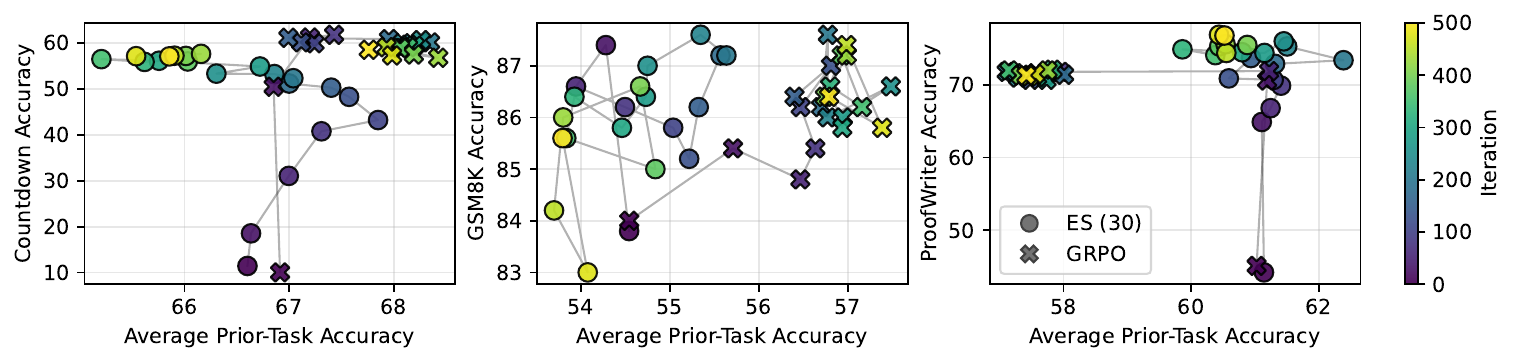}
    \caption{\emph{Target task accuracy vs. average prior task accuracy throughout training.} Colors denote the iteration within the training process. For ES, prior task accuracy exhibits noticeable drift, where accuracy decreases initially, but often recovers later in training. Moreover, forgetting does not arise under all settings. GRPO exhibits less drift in prior task accuracy, but can also lead to severe forgetting in some settings. Detailed results per prior-task are provided in Fig.~\ref{fig:individual_countdown}, Fig.~\ref{fig:individual_gsm8k} and Fig.~\ref{fig:individual_proofwriter}.
    }
    \label{fig:forgetting_tasks}
\end{figure}

\begin{figure}[]
    \centering
    \includegraphics[width=\linewidth, trim=0.2cm 0.2cm 0.2cm 0.2cm, clip]{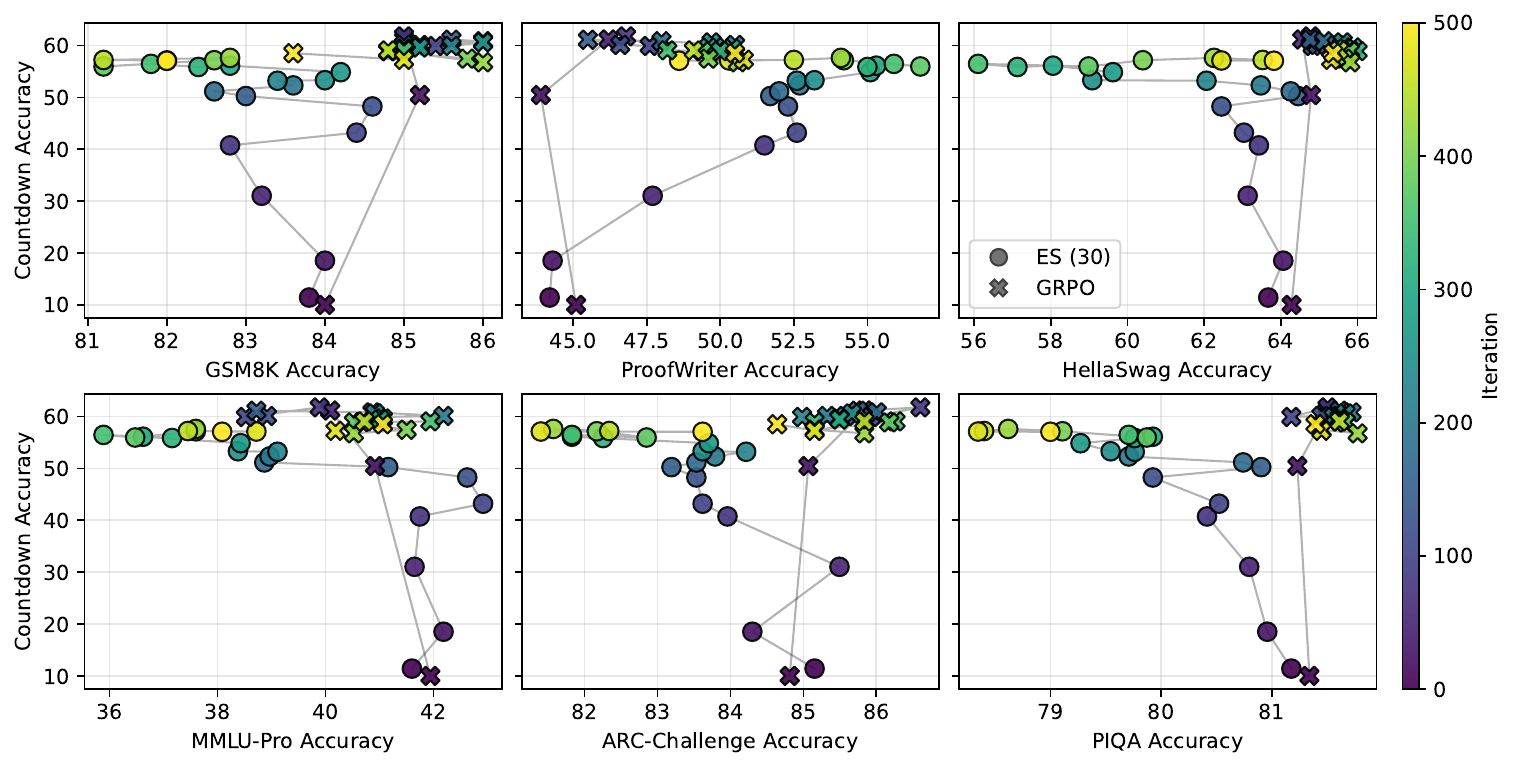}
    \caption{\emph{Individual prior task accuracies training on Countdown as target task.} Colors denote the iteration within the training process. A performance drift rather than irreversible forgetting is observed across multiple tasks. For example, performance on HellaSwag drops by 8\% accuracy over the first 300 iterations, but subsequently recovers to its original level by the final iteration.}
    \label{fig:individual_countdown}
\end{figure}

First, the change in prior task accuracy over the training process is investigated.
Exemplary results for ES and GRPO under the standard experimental conditions (Sec.~\ref{sec:setup}) for the three different target tasks Countdown, GSM8K and ProofWriter are provided in Fig.~\ref{fig:forgetting_tasks}.
Individual results per prior-task for the Countdown experiment are provided in Fig.~\ref{fig:individual_countdown}.
For GSM8K and ProofWriter, they are provided in Fig.~\ref{fig:individual_gsm8k} and Fig.~\ref{fig:individual_proofwriter} in App.~\ref{app:detailed_results}.

The results for Countdown as target task are largely consistent with those reported by \citet{Abdi:26}: ES exhibits prior-task forgetting, whereas GRPO does not.
However, the change in performance for some individual prior tasks (HellaSwag, MMLU-Pro and ARC-Challenge) shows an unexpected pattern.
Performance initially drops, but recovers later in training.
For ProofWriter as prior task, the opposite trend is observed, an initial improvement followed by a return to baseline performance.
Together, these observations suggest that \textbf{ES does not induce irreversible forgetting, but rather a performance drift that is transient}.
A possible explanation, as also suggested by \citet{Hoy:26}, is that ES is free to explore directions in weight space that are weakly constrained by the target task, which may lead to both decreases and increases in performance on unrelated tasks.

The results for GSM8K and ProofWriter in Fig.~\ref{fig:forgetting_tasks} show a different picture.
For those, there is no pronounced prior task forgetting under ES on average over the considered prior tasks.
While for some individual prior-tasks, ES still exhibits a moderate amount of forgetting, there are also some without forgetting.
Furthermore, for ProofWriter as target task, GRPO exhibits considerable forgetting.
Fig.~\ref{fig:individual_proofwriter} shows, that this is mostly due to a strong accuracy decrease on GSM8K as prior task, yet also others (Countdown, MMLU-Pro) show degradation.
Overall, those findings demonstrate that \textbf{forgetting is not a specific failure mode of ES, but can also arise for GRPO.}

While analyzing the training behavior throughout the training process gives insights into the mechanisms of prior-task forgetting, it is hard to compare many different experimental conditions.
The subsequent analyses consider only the change in accuracies between the base model and the final training iteration, allowing systematic comparisons between settings.
Next, the dependence of forgetting during LLM fine-tuning on the target task and model type is analyzed.

\subsection{Dependence of target task and model types on forgetting} \label{sec:forgetting_dependence}

\begin{figure}
    \centering
    \includegraphics[width=\linewidth, trim=0.2cm 0.2cm 0.2cm 0.2cm, clip]{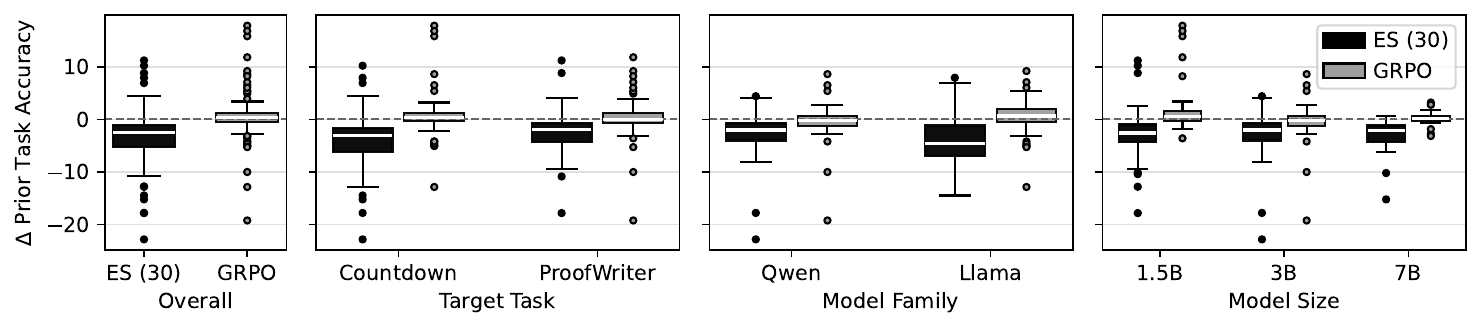}
    \vspace{-0.5cm}
    \caption{\emph{Change in final prior-task accuracy under different target tasks and base models.} Boxplots for the change ($\Delta$) in prior task accuracy for ES and GRPO. Each boxplot aggregates results across all experimental dimensions (target-tasks, prior-tasks, model types) not defined via the x-axis. Target task, model family (restricting aggregation to 3B models) and model size (restricting aggregation to Qwen models) are varied. No significant dependence on varied factors is observed.}
    \label{fig:influence}
    \vspace{-0.1cm}
\end{figure}

Prior work on forgetting in ES and GRPO has focused on a limited selection of target and prior tasks, as well as a single model family and scale \citep{Abdi:26, Hoy:26}, limiting the understanding of how those choices influence this phenomenon. 
Therefore, the target task (Countdown, ProofWriter), model family (Qwen, Llama) and model size (1.5B, 3B, 7B) are varied systematically and evaluated across multiple prior tasks to assess the extent to which forgetting depends on those factors.

The results are summarized in Fig.~\ref{fig:influence}.
ES in the standard setting with a population size of 30 leads to more forgetting than GRPO on average, with a median change in prior accuracy of -2.5 compared to +0.3.
However, both methods exhibit strong positive as well as negative changes in accuracy in some settings.
Forgetting with both ES and GRPO is comparable under Countdown and ProofWriter as target tasks, with more discrepancy observed between the methods for the Countdown task.
Similarly, the discrepancy between ES and GRPO is stronger for Llama models (comparison restricted to the 3B models).
Between models of different model sizes (comparison restricted to the Qwen model family), no significant dependence is observed.

Overall, the consistency of observed trends across model families and sizes suggests that the effect is rather a product of the training dynamics of fine-tuning methods than a property of a specific model.
Therefore, the impact of ES update characteristics on forgetting is investigated next.
As forgetting under fine-tuning with ES appears to be most prominent for the target task Countdown, the remainder of the presented experiments in the main paper focus on this target task.

\begin{table}[b!]
\vspace{-0.1cm}
    \centering
    \caption{\emph{Change in target and prior task accuracies for different ES population sizes.} The target task is Countdown and the average is calculated over accuracies in all prior tasks, thus excluding Countdown. Statistics are computed over three runs; for baseline accuracies see Tab.~\ref{tab:accuracies} in App~\ref{app:detailed_results}. There is less prior task degradation for higher population sizes. \vspace{0.1cm}}
    \label{tab:popsize_ms}
    \footnotesize
    \setlength{\tabcolsep}{4.3pt}
    \begin{tabular}{c|ccccccc|c}
    \toprule
    Method & \hl{Countdown} & GSM8K & ProofWriter & HellaSwag & PIQA & ARC-C & MMLU-Pro & Average \\
    \midrule
    ES (30) & $+45.3_{(0.3)}$ & $-9.5_{(11.5)}$ & $-2.0_{(5.7)}$ & $-8.2_{(9.0)}$ & $-2.8_{(1.0)}$ & $-2.0_{(0.5)}$ & $-6.5_{(2.6)}$ & $-5.2_{(4.7)}$ \\
    ES (128) & $+48.6_{(0.9)}$ & $+0.9_{(2.0)}$ & $-1.4_{(9.3)}$ & $+0.5_{(2.0)}$ & $-0.9_{(0.3)}$ & $-0.4_{(0.4)}$ & $-2.4_{(0.7)}$ & $-0.6_{(1.9)}$ \\
    ES (256) & $+48.4_{(1.1)}$ & $-4.2_{(5.7)}$ & $+6.5_{(3.7)}$ & $+0.1_{(1.7)}$ & $-0.7_{(0.6)}$ & $-0.1_{(0.5)}$ & $-0.0_{(1.7)}$ & $+0.3_{(1.8)}$ \\
    \bottomrule
    \end{tabular}
     
\end{table}

\subsection{The role of ES update characteristics on forgetting} \label{sec:gradient_quality}

\citet{Hoy:26} argued that ES updates decompose into a signal component and a noise component, the latter inducing a random walk in low-curvature directions that are largely irrelevant to target-task performance.
Similar arguments were made by \citet{Liang:26}.
Consistent with this view, \citet{Qiu:25} showed empirically that random walk contributes substantially to ES updates.
Under the theoretical model for ES updates of \citet{Hoy:26}, the expected cumulative weight deviation induced through random walk is given by
\begin{equation} \label{eq:deviation}
    \mathbb{E}\left[ ||\btheta_T - \btheta_0||^2_2 \right] = \frac{\alpha^2 T d}{N} \ ,
\end{equation}
where $\alpha$ is the stepsize, $T$ is the number of iterations, $d$ is the number of weights in the model, and $N$ is the population size.
Under fixed $\alpha$ and $d$, it grows linearly in the number of iterations and shrinks inversely with the population size.
Thus, less random walk behavior and potentially less forgetting is expected under higher population sizes.
This is confirmed by the empirical results (see Tab.~\ref{tab:popsize_ms}) showing that larger population sizes substantially reduce forgetting.
Furthermore, Sec.~\ref{sec:magnitude} empirically validates that Eq.~\eqref{eq:deviation} derived by \citet{Hoy:26} for the theoretical model explains the empirical behavior of ES well, as the update norm is indeed inversely proportional to the population size.
Overall, the results suggest that random walk in weight space is the primary driver of forgetting in ES fine-tuning. 
Accordingly, a method to limit the random walk behavior is introduced next.

\section{Overcoming Forgetting in ES Fine-Tuning} \label{sec:awd}

Given the mechanisms of forgetting outlined in the previous section, an important question is how it can be mitigated.
While increasing the population size substantially reduces forgetting, it incurs significant computational cost.
This section introduces Anchored Weight Decay, a simple mechanism that achieves comparable improvements even at small population sizes.

\paragraph{Reducing the random walk.}
The theoretical analysis of ES optimization dynamics by \citet{Hoy:26} suggested that ES updates induce random walk in directions of the weight space that are weakly coupled to the target task.
Over multiple iterations, random update directions may accumulate and lead to the prior-task performance drift observed in the experiments.
This effect can be counteracted through an explicit constraint that biases the optimization trajectory toward the initial weights $\btheta_0$.
A principled way to formalize such a constraint is through introducing a regularization term into the ES objective, where $l(\cdot)$ is a penalty function and $\lambda > 0$ controls the strength of the constraint:
\begin{equation} \label{eq:objective}
    \max_{\btheta} \mathbb{E}_{\beps \sim \mathcal{N}(\bm{0}, I)} \left[ R(\btheta + \sigma \beps) \right] - \lambda \sum_{i=1}^d l(\btheta - \btheta_{0})_i \ .
\end{equation}
This formulation allows for general element-wise penalty functions $l(\cdot)$.
For example, $l(x) = \tfrac{1}{2} x^2$ yields an $\ell_2$ penalty inducing smooth shrinkage toward $\btheta_0$, while $l(x) = |x|$ corresponds to an $\ell_1$ penalty, promoting sparsity in the weight difference $\btheta - \btheta_0$.

Notably, the objective in Eq.~\eqref{eq:objective} is related to Elastic Weight Consolidation (EWC) \citep{Kirkpatrick:17}, which considers a quadratic penalty function and weights the regularization of individual parameters by the Fisher information matrix. 
In contrast, Eq.~\eqref{eq:objective} assigns equal weight to all parameters. 
This reflects the absence of task-specific importance information in the present setting, as LLMs may be used for a huge variety of prior tasks.
Empirical results from prior work \citep{Li:18} further suggest that such uniform weighting can perform comparably to Fisher-based approaches in practice.
A detailed discussion of the relationship to prior work on regularization based mitigation of prior task forgetting is given in App.~\ref{app:ewc}.

\paragraph{Anchored Weight Decay.}
For gradient-based fine-tuning, regularized objectives of the form as in Eq.~\eqref{eq:objective} are usually implemented by adding the penalty term to the loss calculation for backpropagation.
That approach is not applicable to ES.
However, $\ell_1/\ell_2$ regularization can equivalently be applied as weight decay directly in the update rule of the weights \citep{Hanson:88, Loshchilov:18}.
Weight decay was introduced with the explicit aim to keep the absolute values of weights close to zero. 
To limit random walk behavior and thus forgetting, those values instead need to be kept close to the initial weights $\btheta_0$. 
The resulting method is called \emph{Anchored Weight Decay} (AWD). 
It extends the ES update rule (Eq.~\eqref{eq:es}) with a decay on the updated parameters:
\begin{equation} \label{eq:awd}
    \btheta_{t} = \textcolor{gray}{\underbrace{\btheta_{t - 1} + \alpha \cdot \frac{1}{N} \sum_{n=1}^{N} \hat{r}_n \beps_n}_{\text{ES Update (Eq.~\eqref{eq:es})}}} \ - \ \alpha \lambda l'\Big(\textcolor{gray}{\underbrace{\btheta_{t - 1} + \alpha \cdot \frac{1}{N} \sum_{n=1}^{N} \hat{r}_n \beps_n}_{\text{ES Update (Eq.~\eqref{eq:es})}}} \ - \ \btheta_0\Big) \ .
\end{equation}
The full ES algorithm with addition of AWD is provided in Algorithm~\ref{alg:basic_es}.
Whether or not to couple the weight decay term with the learning rate is a design choice that may be investigated in more detail in future work.
The ES implementation of \citet{Qiu:25} utilized a fixed $\alpha$ across iterations, thus $\alpha$ could be absorbed into $\lambda$. 
Yet under adaptive $\alpha$, it may be advantageous to decouple them, as done by \citet{Loshchilov:18}.

AWD can be understood as a proximal optimization step or a form of trust region method that counteracts the accumulation of task-irrelevant drift in weight space. 
It retains the simplicity and scalability of ES, requires no additional forward passes, and provides a flexible mechanism to control the geometry of the constraint via the choice of penalty function $l(\cdot)$.

\paragraph{Implementation.}
Compared to vanilla ES as in \citet{Qiu:25}, AWD requires access to the reference parameters $\btheta_0$ to compute the decay term.
Keeping a copy of $\btheta_0$ in VRAM during population evaluation is often impractical, because GPU memory is better utilized for batched sequence generation and reward computation.
Instead, $\btheta_0$ is stored in contiguous pinned RAM and streamed layer-wise to the GPU on demand during the weight update in Eq.~\eqref{eq:awd}.
The additional overhead introduced by AWD is negligible in practice, as data transfer is only needed once every iteration and does not interfere with the compute- and bandwidth-intensive evaluation phase.

\begin{algorithm}[h!]
\caption{Basic ES Algorithm \textcolor{magenta}{with Anchored Weight Decay (AWD)}}
\label{alg:basic_es}
\begin{algorithmic}[1]
\Require Pretrained LLM with initial parameters $\btheta_0$, reward function $R(\cdot)$, total iterations $T$, population size $N$, noise scale $\sigma$, learning rate $\alpha$, \textcolor{magenta}{penalty factor $\lambda$, penalty function $l(\cdot)$}
\For{$t = 1$ to $T$} \Comment{outer ES iteration over generations of populations}
    \For{$n = 1$ to $N$} \Comment{inner ES iteration over population members}
        \State Sample noise $\beps_n \sim \mathcal{N}(\bm{0}, I)$
        \State $r_n \gets R(\btheta_{t-1} + \sigma \beps_n)$ \Comment{evaluate perturbed model on train samples}
    \EndFor
    \State $\hat{r}_n \gets$ normalize($r_n$) \Comment{z-scaling rewards over the population}
    \State $\btheta_t \gets \btheta_{t-1} + \alpha \cdot \frac{1}{N} \sum_{n=1}^{N} \hat{r}_n \beps_n$ \Comment{calculate standard ES update}
    \State $\textcolor{magenta}{\btheta_t \gets \btheta_{t} - \alpha \lambda l'(\btheta_{t} - \btheta_0)}$ \Comment{decay towards initial parameters}
\EndFor
\end{algorithmic}
\end{algorithm}

\begin{table}[b!]
    \centering
    \vspace{-0.3cm}
    \footnotesize
    \setlength{\tabcolsep}{2.2pt}
    \caption{\emph{Change in target and prior task accuracies using AWD with $\ell_1/\ell_2$ decay.} The target task is Countdown and the average is calculated over accuracies in all prior tasks, thus excluding Countdown. Statistics are computed over three runs; for baseline accuracies see Tab.~\ref{tab:accuracies} in App~\ref{app:detailed_results}. AWD effectively reduces forgetting for both $\ell_1$ ($\lambda = 0.01$) and $\ell_2$ ($\lambda = 10.0$) decay. \vspace{0.1cm}}
    \label{tab:awd}
    \begin{tabular}{c|ccccccc|c}
    \toprule
    Run & \hl{Countdown} & GSM8K & ProofWriter & HellaSwag & PIQA & ARC-C & MMLU-Pro & Average \\
    \midrule
    ES (30) & $+45.3_{(0.3)}$ & $-9.5_{(11.5)}$ & $-2.0_{(5.7)}$ & $-8.2_{(9.0)}$ & $-2.8_{(1.0)}$ & $-2.0_{(0.5)}$ & $-6.5_{(2.6)}$ & $-5.2_{(4.7)}$ \\
    ES (30) + AWD ($\ell_1$) & $+44.3_{(1.0)}$ & $-0.1_{(1.2)}$ & $+3.7_{(2.1)}$ & $-2.1_{(0.9)}$ & $-1.9_{(1.2)}$ & $-1.2_{(1.0)}$ & $-3.5_{(2.0)}$ & $-0.9_{(0.6)}$ \\
    ES (30) + AWD ($\ell_2$) & $+43.8_{(0.7)}$ & $-2.5_{(5.0)}$ & $+1.0_{(6.8)}$ & $-0.7_{(1.6)}$ & $-0.9_{(0.2)}$ & $-0.5_{(1.1)}$ & $-2.6_{(0.9)}$ & $-1.0_{(2.4)}$ \\
    \bottomrule
    \end{tabular}
    \vspace{-0.1cm}
\end{table}

\begin{figure}[b!]
    \vspace{-0.1cm}
    \includegraphics[width=\linewidth, trim=0.2cm 0.2cm 0.2cm 0.2cm, clip]{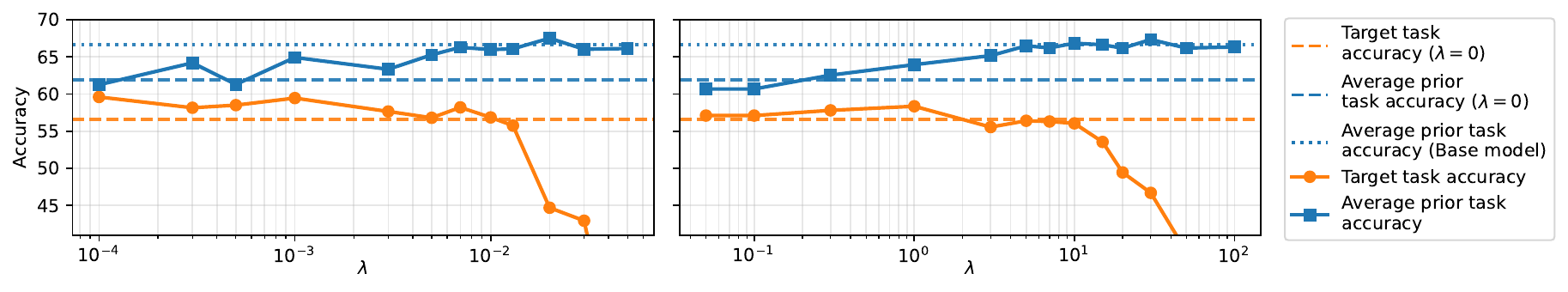}
    \begin{subfigure}{0.45\linewidth}
    \vspace{-0.2cm}
        \subcaption{$\ell_1$ penalty function}
    \end{subfigure}
    \begin{subfigure}{0.33\linewidth}
    \vspace{-0.2cm}
        \subcaption{$\ell_2$ penalty function}
    \end{subfigure}
    \vspace{-0.15cm}
    \caption{\emph{Effect of the AWD penalty factor $\lambda$ on target- and prior-task accuracy for (a) $\ell_1$ and (b) $\ell_2$ penalty function.} Target task is Countdown, dashed line shows target- and prior task accuracies for ES without AWD, dotted line shows prior-task accuracy of the base model. For both penalty functions, there is an intermediate range of $\lambda$ that preserves target-task performance while mitigating forgetting.}
    \vspace{-0.3cm}
    \label{fig:ablation_lambda}
\end{figure}

The decay under $\ell_1$ penalty is implemented as proximal step, thus $l'(x) = \min(\alpha \lambda, |x|) \cdot \text{sign}(x)$ instead of $\alpha \lambda \cdot \text{sign}(x)$.
However, the proximal step usually leads to better solutions of Eq.~\eqref{eq:objective}, as it sets the weight difference exactly to zero (induces sparsity) if it is smaller than $\alpha \lambda$.

\paragraph{On the choice of penalty function $l(\cdot)$.}
The effectiveness of applying AWD with $\ell_1$ and $\ell_2$ penalty to ES is empirically evaluated. 
Fig.~\ref{fig:main} (left) illustrates the difference in optimization behavior between vanilla ES and ES+AWD ($\ell_2$), showing that AWD effectively mitigates prior task forgetting. 
Individual prior-task results are reported in Fig.~\ref{fig:individual_awd} in the Appendix.
A comparison between $\ell_1$ and $\ell_2$ penalties is provided in Tab.~\ref{tab:awd}; both penalty types were found to lead to similar improvements.
Additional results on other combinations of target dataset and model type are provided in App.~\ref{app:detailed_awd}, further demonstrating the effectiveness of AWD under both $\ell_1$ and $\ell_2$ penalties.

\paragraph{On the choice of penalty factor $\lambda$.}
Furthermore, the sensitivity of both target-task performance and prior-task forgetting w.r.t. the penalty coefficient $\lambda$ is analyzed. 
As shown in Fig.~\ref{fig:ablation_lambda}, small values of $\lambda$ yield little reduction in prior task forgetting compared to ES without AWD. 
However, target-task performance appears to slightly improve under $\ell_1$ regularization at low $\lambda$. 
For both penalty types, there exists a stable range of $\lambda$ where AWD mostly preserves target performance while substantially reducing forgetting.
Beyond this range, prior-task performance remains stable, but target-task performance degrades. 
The sharp drop at large $\lambda$ suggests a practical tuning strategy: start with a high $\lambda$ and decrease it until no performance drop is observed relative to ES without AWD.

AWD was introduced to mitigate forgetting caused by random walk in weight space.
While the results confirm its effectiveness, it remains unclear whether this improvement stems from reduced weight drift, which is investigated in the next section.

\section{Analyzing Model Changes under ES with Random Walk Mitigation}
\label{sec:magnitude}

According to Eq.~\eqref{eq:deviation}, increasing the population size reduces the expected norm of the cumulative weight deviation, induced by random walk in the directions weakly constrained by the target task.
AWD was introduced with the same objective of reducing such random walk.
Therefore, the effect of AWD as well as large population sizes on the weight updates is investigated in this section.
Complementarily, an analysis of KL-divergence across tasks is provided, assessing how differences in the magnitude of weight changes translate into changes in the output distribution.

\begin{wrapfigure}{r}{0.45\textwidth}
    \centering
    \vspace{-0.28cm}
    \includegraphics[width=\linewidth, trim=0.2cm 0.2cm 0.2cm 0.2cm, clip]{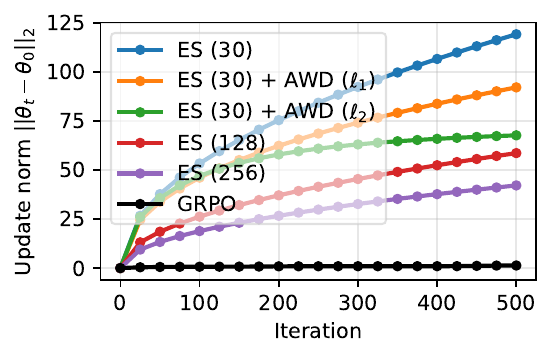}
    \caption{\emph{Norm $||\btheta_t - \btheta_0||_2$ of weight updates.} Larger ES populations and AWD reduce update norms. ES + AWD ($\ell_2$) converges to levels similar to ES with population size 128. GRPO updates remain over an order of magnitude smaller.}
    \label{fig:weight_change}
    \vspace{-0.4cm}
\end{wrapfigure}

\paragraph{Weight updates.}
Fig.~\ref{fig:weight_change} shows the euclidean norm of changes in the weights $||\btheta_t - \btheta_0||_2$ for each iteration $t$ throughout training.
Consistent with \citet{Abdi:26}, ES with a population size of 30 produces updates (cumulative weight changes) with norms roughly two orders of magnitude larger than those for GRPO.
Increasing the population size to 128 reduces the magnitude of the update norm by about half; however, even at a population size of 256, the update norm remains more than an order of magnitude higher.

In contrast, applying AWD to ES with a population size of 30 substantially reduces the update norm, demonstrating its ability to limit random walk in weight space.
For $\ell_2$ updates, the norm initially matches the case without AWD, but eventually converges to a level comparable to that observed with a population size of 128.
Nevertheless, a substantial gap to GRPO remains, raising the question of whether the high-norm updates induced by ES lead to meaningful shifts in LLM behavior.

\begin{figure}
    \centering
    \includegraphics[width=\linewidth, trim=0.2cm 0.2cm 0.2cm 0.2cm, clip]{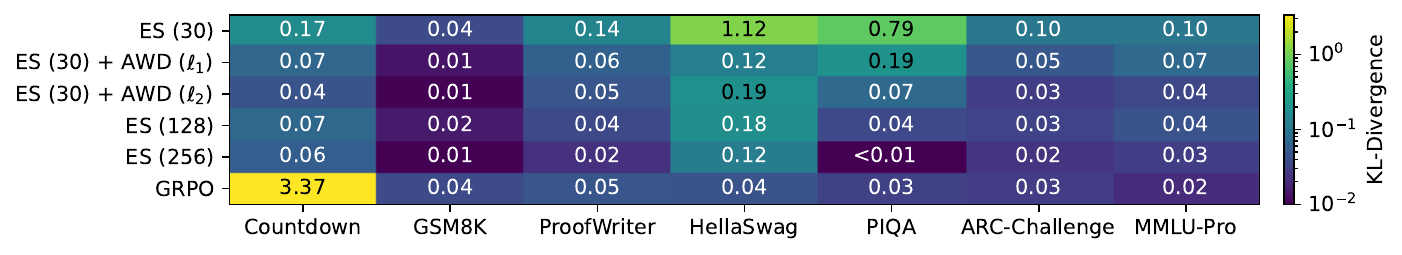}
    \caption{\emph{KL-divergence} $\text{KL}(p_{\btheta_0} || p_{\btheta_T})$ \emph{between base and final model trained on Countdown.} ES with a small population size (30) causes larger shifts on prior tasks than GRPO; increasing population size or adding AWD mostly closes this gap. On the target task Countdown, GRPO shows very high divergence while it remains low for all ES variants, despite all having similar target task accuracy.}
    \label{fig:kl_divergence}
\end{figure}

\paragraph{Distributional shift.}
Fig.~\ref{fig:kl_divergence} compares the KL-divergence $\text{KL}(p_{\btheta_0} || p_{\btheta_T})$ between the base model and the final model after training on the Countdown task.
Consistent with the trends observed in Fig.~\ref{fig:weight_change}, ES with a small population size (30) induces substantially larger distributional shifts on prior tasks than GRPO, most prominently on HellaSwag and PIQA.
Increasing the population size systematically reduces this divergence, indicating that larger populations mitigate the extent to which ES deviates from the base model, roughly matching the level of KL-divergence exhibited on prior tasks by GRPO.
A similar effect is achieved by introducing AWD, especially under $\ell_2$ penalty function.
Thus, even though the update norm is still much higher than for GRPO, their induced distributional shift as measured by the KL-divergence is comparable.

A markedly different pattern emerges on the target task. 
GRPO exhibits much higher KL-divergence on Countdown, whereas all ES variants maintain comparatively low divergence despite achieving the same accuracy.
While \citet{Shenfeld:25} reported that RL methods such as GRPO adapt to target tasks with much lower KL divergence than supervised fine-tuning, the results show that ES reduces divergence on the target task even further. 
This observation matches the studies in \citet{Qiu:25}.

Overall, the results show that while higher population sizes and AWD substantially reduce the magnitude of the weight change, there still remains a large gap between GRPO and ES.
However, GRPO and ES with AWD or high population size lead to similar distributional shift on prior tasks.
Furthermore, GRPO induces much greater distributional shift on the target task, albeit having similar accuracy.
The conclusion is that it is the randomness of the drift unconstrained by the target task that leads to prior task forgetting for ES, rather than the magnitude of updates alone.

\section{Discussion} \label{sec:discussion}

This paper analyzed prior-task forgetting for ES fine-tuning and introduced AWD to mitigate it.
The results show that the extent of such forgetting is mostly driven by random walk behavior.
Furthermore, it was found that forgetting is not limited to ES, but can also arise for GRPO.
For settings where the issue arises, increasing the population size is an effective mitigation, albeit at increased computational expense.
The proposed AWD method is equally effective even with small population sizes, and thus more efficient to employ in practice.
Finally, both AWD and higher population sizes decrease the norm of weight updates.
While they are still an order of magnitude higher than for GRPO, they induce similar distributional shift on the prior tasks, suggesting that the norm of weight updates alone does not determine forgetting.

\paragraph{Limitations.}
AWD incurs a small additional compute cost for the penalty term. 
However, this overhead is less significant with larger population sizes, because it is applied only once per iteration.
Unlike standard ES, AWD requires access to the original weights each iteration, increasing memory usage when stored on the GPU or runtime when streamed from RAM. 
The experiments were performed using the second option, where using AWD increased the runtime about 1-2\%.

Moreover, this study focused on prior-task forgetting in verifiable domains. 
Extending this analysis to alignment and safety degradation---and testing whether AWD preserves these properties, not just prior task accuracy---is an important direction for future work.

\paragraph{The blessing and curse of random walks.}
The effectiveness of AWD and larger population sizes point to random walk in directions of weight space weakly coupled to the target task as the primary driver of forgetting.
This random walk reflects a fundamental difference between ES and RL: as noted by \citet{Hoy:26}, when the reward signal diminishes, RL remains stationary in weight space while ES continues to explore. 
While AWD restricts this behavior to preserve prior-task performance, the same random walk may allow ES to escape local optima that RL stays trapped in, which is an potential benefit worth investigating empirically in future work.

\paragraph{The choice of penalty function.}
The results of the comparison between $\ell_1$ and $\ell_2$ penalty functions in Sec.~\ref{sec:awd} showed that $\ell_2$ is slightly more robust in practice and led to a smoother decline of target task performance when it was set too high.
However, induced sparsity of weight updates under $\ell_1$ could potentially be used to improve target task performance.
Setting $\lambda$ smaller than the critical magnitude to retain prior-task performance systematically improved target task performance. 

While widely used, $\ell_1$ and $\ell_2$ are only two instances of penalty functions.
For instance, combining the two as the Huber penalty function \citep{Huber:64} could combine the benefits of both.
Future work should investigate the effect of different types of penalty functions on prior-task forgetting under ES.

\paragraph{A scalable path to lifelong learning.}
Given the results on mitigating prior-task forgetting, ES is well suited for developing agentic systems that continually learn during deployment.
Because ES relies only on forward passes, it can fully exploit throughput-optimized hardware without a separate training stack, making it a compelling method for scalable, continually improving systems.

\section{Conclusion}

Previous works pointed out that ES fine-tuning, while competitive for target task performance, suffers from prior task forgetting \citep{Abdi:26, Hoy:26}.
This work shows that prior task forgetting during ES fine-tuning is largely avoidable, either intrinsically through larger population sizes or through adding AWD to the base algorithm.
This positions ES well for continual learning systems deployed in the real world, where the simplicity and scalability of ES pose unique advantages.

\begin{ack}
We thank Akshat Gupta and Immanuel Abdi for helpful discussions on reproducing their results.%
\end{ack}

\bibliography{literature}
\bibliographystyle{plainnat}

\clearpage

\appendix

\section{Broader Impact} \label{app:impact}

This work contributes to improving the reliability of LLM fine-tuning by addressing prior-task degradation during post-training.
It is shown that higher population sizes, as well as the introduced method AWD are effective in mitigating this issue.
This has positive implications for real-world deployment, where models must adapt to new tasks without losing existing capabilities.
Reducing unintended degradation also lowers the need for repeated retraining, potentially decreasing computational costs and environmental impact.

However, enabling more effective continual learning may also accelerate the deployment of increasingly autonomous systems that adapt over time. 
This raises concerns around monitoring, evaluation, and control, as models could evolve in unintended ways if not properly constrained. 
While AWD has been found effective to mitigate performance drift relative to the initial model, its ability to preserve alignment or safety properties during fine-tuning has not been tested yet, which is an important direction for future work.

Finally, the  findings of this paper highlight that fine-tuning dynamics play a central role in distribution shift. 
This underscores the importance of developing robust evaluation protocols and safeguards under continuous model updates.

\section{Background on Group Relative Policy Optimization} \label{app:grpo}

Group Relative Policy Optimization (GRPO) \citep{Shao:24} is an RL–based method for fine-tuning LLMs. It is a Proximal Policy Optimization (PPO) \citep{Schulman:17} variant that avoids the need for an explicit value function.
Given an input sequence $x$, GRPO samples a group of response sequences $\{y_k\}_{k=1}^K$ from the current policy (LLM) $p_{\btheta}$ and computes rewards $R(y_k)$.
Instead of estimating advantages via a learned critic, GRPO normalizes rewards within the sampled group to obtain advantages $\hat{A}_k = \tfrac{R(y_k) - \mu_K}{\sigma_K}$, where $\mu_K$ and $\sigma_K$ are the mean and standard deviation over the sampled group of responses.
The policy is then updated using a clipped surrogate objective as in PPO \citep{Schulman:17}:
\begin{equation} \label{eq:grpo}
    J_{\text{GRPO}}(\btheta) = \mathbb{E} \left[ \frac{1}{K} \sum_{k=1}^K \min \left(\rho_{k}(\btheta) \hat{A}_k, \ \text{clip}(\rho_{k}(\btheta), 1 - \epsilon, 1 + \epsilon) \hat{A}_k \right) \right] \ ,
\end{equation}
where $\rho_{k}(\btheta) = \frac{p_{\btheta}(y_k \mid x)}{p_{\btheta_{\text{old}}}(y_k \mid x)}$ is the likelihood ratio of the output sequence between the current model and a previous version of the model.
Usually, Eq.~\eqref{eq:grpo} is extended by a KL-divergence penalty term between the current policy and the original policy, acting as regularization.
The weights $\btheta$ of the policy are then updated via backpropagation to maximize the objective.
GRPO preserves the effectiveness of PPO while being simpler and more efficient due to the removal of the value model.
Therefore, it has emerged as a strong default for LLM post-training, particularly in reasoning tasks.
However, implementations of RL-based pipelines are complex, requiring separate systems for sampling output trajectories \citep{Kwon:23, Zheng:24} and for training the model via backpropagation \citep{Zhao:23, Shoeybi:19}, increasing engineering overhead and limiting scalability.
Scaling these systems to multiple processing units or even compute nodes is a major engineering challenge, as gradients and optimizer states need to be synchronized between different processing units.
In contrast, ES requires just a single system for sampling output trajectories and scales without much engineering effort as solely scalar information about seeds and rewards needs to be exchanged between processing units.

\section{Connection to Regularization-Based Mitigation Methods for Forgetting} \label{app:ewc}

The works of \citet{Kirkpatrick:17} and \citet{Zenke:17} have demonstrated that regularizing changes to the weights when training on a new task is effective to mitigate prior task forgetting of neural networks.
In the following, their connection to AWD is highlighted, and it is discussed how AWD can be interpreted through the lens of their papers motivations.

\paragraph{Connection to \citet{Kirkpatrick:17}.}
Elastic Weight Consolidation (EWC) \citep{Kirkpatrick:17} considers a quadratic penalty towards the original weights $\btheta_0$, that is weighted by their importance on prior tasks.
EWC is motivated through the Bayesian framework:
The relevance of $\btheta$ w.r.t the training data $\mathcal{D}$ is given by its posterior probability $p(\btheta \mid \mathcal{D}) = \tfrac{p(\mathcal{D} \mid \btheta) p(\btheta)}{p(\mathcal{D})}$ (Bayes rule).
Importantly, the log probability of the data given the weights $\log p(\mathcal{D} \mid \btheta)$, their log-likelihood, is nothing else than the negative of the standard cross entropy loss function $- \mathcal{L}(\btheta)$.
Assuming the data can be split into two independent parts that define tasks $A$ and $B$ and applying the logarithm to Bayes rule, allows to rearrange to
\begin{equation}
    \log p(\btheta \mid \mathcal{D}) = \log p(\mathcal{D}_B \mid \btheta) + \log p(\btheta \mid \mathcal{D}_A) - \log p(\mathcal{D}_B) \ ,
\end{equation}
Thus all information about task $A$ is provided by the posterior distribution $p(\btheta \mid \mathcal{D}_A)$.
It thus provides information, which weights were important for task $A$, yet the posterior is generally intractable to compute exactly.
EWC follows the ideas of the Laplace posterior approximation in assuming a Gaussian centered around the original weights $\btheta_0$, i.e. weights that have been optimized on task $A$.
The precision of the Gaussian considers a diagonal approximation for tractability (the covariances between weights are neglected), which is given by the diagonal of the Fisher information matrix $F$ for the task $A$, thus $F_{i} = - \mathbb{E} \left[ \frac{\partial^2}{\partial \theta_i^2} \log p(\mathcal{D}_A \mid \btheta) \mid \btheta \right]$.
Thus, they implement the loss function
\begin{equation} \label{eq:ewc}
    \mathcal{L}(\btheta) = \mathcal{L}_B(\btheta) + \frac{\lambda}{2} \sum_{i=1}^d F_{i} (\btheta_i - \btheta_{0,i})^2 \ .
\end{equation}
This is related to the definition of the regularized objective function in Eq.~\eqref{eq:objective}, where EWC considers the $\ell_2$ penalty function.
While for gradient based methods, it is applicable to compute the regularization term as part of the loss function and automatically change the update of weights through backpropagation, for ES this is not applicable and it is necessary to directly implement AWD in the weight update equation.
Furthermore, in the setting considered in this paper, there is no fixed prior task $A$, but countless possible prior tasks that the LLM could be used for.
Considering the Bayesian formulation that motivated EWC, AWD models the posterior $p(\btheta \mid \mathcal{D}_{A})$ with a uniform distribution, following the maximum entropy principle, arriving at a uniform weighting.
If there is a certain set of tasks that one cares about, AWD could be further improved by modeling $F$ as done by EWC.

\paragraph{Connection to \citet{Zenke:17}.}
The regularization approach of \citet{Zenke:17} arrives at a similar loss term as EWC, yet with different motivation and interpretation of the weighting term $\Omega_i$, and is given by 
\begin{equation} \label{eq:synaptic}
    \mathcal{L}(\theta) = \mathcal{L}_B(\btheta) + \frac{\lambda}{2} \sum_{i=1}^d \Omega_{i} (\btheta_i - \btheta_{0,i})^2 \ ,
\end{equation}
where the formulation compared to the original paper is adapted to highlight the similarity to EWC and AWD.
\citet{Zenke:17} consider a sequence of tasks, which is presented here in the special case of a two task setting $A$ followed by $B$.
Their weighting term is given by $\Omega_{i} = \tfrac{\omega^A_i}{\Delta\btheta^A_i}$, where $\omega^A_i$ is the parameter-specific contribution to the change in loss for task $A$ and $\Delta\btheta^A_i$ denotes how much the weights changed for learning task $A$.
It thus takes the whole optimization trajectory on task $A$ into account for determining the weighting, not only the local sensitivity around the final model weights as in EWC.
In practice, $\omega^A_i$ is approximated as the running sum of the product between the current gradient $\frac{\partial \mathcal{L}}{\partial \btheta_i}$ and the update to the parameter $\frac{\partial \btheta_i}{\partial t}$.

The connection to AWD is similar as between EWC and AWD, yet the interpretation of using a uniform weighting for $\Omega_{i}$ differs.
Intuitively, having uniform $\Omega_{i}$ means that every weight contributed equivalently to minimizing the loss on all prior tasks.
Again, if there are some tasks that one cares about to not loose performance, a non-uniform weighting for AWD can be introduced as in \citet{Zenke:17} by estimating $\Omega_{i}$ on those tasks.

\section{Detailed Description of Experimental Setup} \label{app:evaluation}

Detailed information on the hyperparameters of ES and GRPO, on the models used for evaluation and descriptions of the considered tasks, including examples, answer format, and extraction logic is provided in the following.

\subsection{Hyperparameters for ES and GRPO} \label{app:hyperparameters}

\paragraph{Tuning strategy.}
For both ES and GRPO, an initial sweep of the most important hyperparameters on the Countdown task for the Qwen2.5 3B Instruct model was conducted.
The parameters selected in \citet{Qiu:25} and \citet{Abdi:26} proved to be very effective.
Then, these parameters were tested on the other target tasks with the same model, which led to good results, with comparable performance gains for both ES and GRPO.
A small additional sweep was performed to confirm that these parameters were competitive, which proved to be the case.
While the target task performance is not the main focus of our study, as this paper mostly studies prior-task performance, it was nevertheless made sure that good models are obtained for a fair comparison.

\paragraph{ES.}
The main hyperparameters selected for ES are as follows: $\sigma=0.001$, $\alpha=\sigma/2$, population size = $30$ (except where otherwise specified by experiments), $500$ training iterations.

\paragraph{GRPO.}
The main hyperparameters selected for GRPO are as follows: lr = 1e-6, batch size = $32$, KL-loss coefficient = $0.001$ (applied to loss, not reward), entropy-loss coefficient = $0$, number of rollouts = $30$, $500$ total epochs.

\subsection{Models}

\paragraph{Qwen-2.5 Instruct} is distributed under the Qwen Research Licence Agreement. \footnote{\url{https://huggingface.co/Qwen/Qwen2.5-3B-Instruct/blob/main/LICENSE}}
The official weights, tokenizer, and chat template as provided through HuggingFace \texttt{transformers} are used.

\paragraph{Llama-3.2 Instruct} is distributed under the Llama 3.2 Community License Agreement. \footnote{\url{https://huggingface.co/meta-llama/Llama-3.2-1B/blob/main/LICENSE.txt}}
The official weights, tokenizer, and chat template as provided through HuggingFace \texttt{transformers} are used.

\subsection{Tasks}
\label{app:tasks}

\paragraph{Countdown} 
considers arithmetic reasoning, where the model must combine a set of given numbers using basic operations to reach a target value.

Published under Apache License 2.0 in the official codebase of \citet{Gandhi:24}.

\begin{tcolorbox}[
  title=Countdown Example,
  colbacktitle=lightgray,
  coltitle=black
]
Using the numbers [44, 19, 35], create an equation that equals 98.
\end{tcolorbox}

\begin{table}[h!]
\centering
\small
\begin{tabular}{p{0.25\textwidth} p{0.69\textwidth}}
\toprule
    Answer format & Extraction logic \\
    \midrule
    \texttt{<answer>...</answer>} & For correctness, all \texttt{<answer>...</answer>} blocks are extracted and only the last one is used; if none exists, reward is zero. Within that block, the text must be non-empty and consist only of \texttt{[0-9+\-*/() ]}, otherwise the reward is zero. Then all integer substrings are extracted, and their multiset must exactly match the provided numbers, i.e.\ every number must appear exactly once. Finally, the expression is evaluated, and the reward is one only if the result matches the target within a tolerance of \texttt{1e-5}; any evaluation error gives zero reward.\\
    \bottomrule
\end{tabular}
\end{table}

\paragraph{GSM8K}
focuses on grade-school math word problems that require multi-step numerical reasoning and understanding of natural language.

Published under MIT license in the official codebase of \citep{Cobbe:21}.

\begin{tcolorbox}[
  title=Example,
  colbacktitle=lightgray,
  coltitle=black
]
Janet’s ducks lay 16 eggs per day. She eats three for breakfast every morning and bakes muffins for her friends every day with four. She sells the remainder at the farmers' market daily for \$2 per fresh duck egg. How much in dollars does she make every day at the farmers' market?
\end{tcolorbox}

\begin{table}[h!]
\centering
\small
\begin{tabular}{p{0.25\textwidth} p{0.69\textwidth}}
\toprule
    Answer format & Extraction logic \\
    \midrule
    \texttt{\#\#\#\# <number>} & Only the last 4096 characters of the response are searched; if \texttt{</think>} appears there, only the text after the last such tag is used. Extraction then tries, in order: last \texttt{\#\#\#\# ...} match, last \texttt{\textbackslash boxed\{...\}} match, last \texttt{<answer>...</answer>} match, otherwise the full search region. The extracted text is parsed numerically by stripping commas, \texttt{\$}, and \texttt{\%}; if it is an exact fraction \texttt{a/b}, that fraction is evaluated, otherwise the first numeric substring is used. Reward is zero if no number can be parsed or if it does not match the target within a tolerance of \texttt{1e-6}. \\
    \bottomrule
\end{tabular}
\end{table}

\paragraph{ProofWriter}
evaluates logical reasoning, requiring the model to derive conclusions from a set of facts and rules using formal inference.

There was no license information provided in the paper, nor for the HuggingFace dataset source. \footnote{\url{https://huggingface.co/datasets/tasksource/proofwriter}}

\begin{tcolorbox}[
  title=Example,
  colbacktitle=lightgray,
  coltitle=black
]
Theory: The dog is not blue. If someone is green and not cold, then they are round.\\
Question: The dog is not blue.
\end{tcolorbox}

\begin{table}[h!]
\centering
\small
\begin{tabular}{p{0.25\textwidth} p{0.69\textwidth}}
\toprule
    Answer format & Extraction logic \\
    \midrule
    \texttt{Answer: <True/False/Unknown>} & Extraction tries, in order: the last multiline match of \texttt{Answer: <word>} anywhere in the response; otherwise the last non-empty line, matched leniently as optional \texttt{Final answer} or \texttt{Answer} followed by a word with optional trailing punctuation; otherwise the last occurrence anywhere of \texttt{true}, \texttt{false}, or \texttt{unknown}; otherwise the whole response is canonicalized. Canonicalization trims whitespace, strips trailing punctuation \texttt{.!?,:;}, lowercases, and maps only \texttt{true/false/unknown} to the valid labels. The reward is zero if canonicalization fails or the label does not exactly match the target. \\
    \bottomrule
\end{tabular}
\end{table}

\paragraph{HellaSwag}
 focuses on commonsense reasoning, where the model selects the most plausible continuation of a given situation.

Published under MIT license in the official codebase of \citet{Zellers:19}.

\begin{tcolorbox}[
  title=Example,
  colbacktitle=lightgray,
  coltitle=black
]
Activity: Roof shingle removal\\
Context: A man is sitting on a roof. he\\
Endings:\\
0. is using wrap to wrap a pair of skis.\\
1. is ripping level tiles off.\\
2. is holding a Rubik's Cube.\\
3. starts pulling up roofing on a roof.\\
\end{tcolorbox}

\begin{table}[h!]
\centering
\small
\begin{tabular}{p{0.25\textwidth} p{0.69\textwidth}}
\toprule
    Answer format & Extraction logic \\
    \midrule
    <\texttt{0}/\texttt{1}/\texttt{2}/\texttt{3}> & The strict format is satisfied only if the entire output is exactly one digit in \texttt{0--3} up to surrounding whitespace. For answer extraction, however, the first occurrence of any character in \texttt{[0-3]} anywhere in the output is used. There are no further fallbacks. The reward is zero if no such digit is found or if it does not equal the gold label. \\
    \bottomrule
\end{tabular}
\end{table}

\paragraph{PIQA}
evaluates physical commonsense understanding, requiring the model to choose solutions that are feasible in real-world scenarios.

Published under Academic Free License v.3.0 in the official codebase of \citet{Bisk:20}.

\begin{tcolorbox}[
  title=Example,
  colbacktitle=lightgray,
  coltitle=black
]
Goal: How do I ready a guinea pig cage for its new occupants?\\
1. Provide the guinea pig with a cage full of a few inches of bedding made of ripped paper strips, you will also need to supply it with a water bottle and a food dish.\\
2. Provide the guinea pig with a cage full of a few inches of bedding made of ripped jeans material, you will also need to supply it with a water bottle and a food dish.
\end{tcolorbox}

\begin{table}[h!]
\centering
\small
\begin{tabular}{p{0.25\textwidth} p{0.69\textwidth}}
\toprule
    Answer format & Extraction logic \\
    \midrule
    <\texttt{1}/\texttt{2}> & The strict format is satisfied only if the entire output is exactly \texttt{1} or \texttt{2} up to surrounding whitespace. For answer extraction, the first standalone \texttt{1} or \texttt{2} found via word boundaries is used without further fallbacks. The reward is zero if no such label is found or if it does not equal the gold label. \\
    \bottomrule
\end{tabular}
\end{table}

\paragraph{ARC-Challenge}
consists of difficult science questions that test reasoning and knowledge beyond simple retrieval.

Listed as CC BY-SA 4.0 according to HuggingFace dataset source. \footnote{\url{https://huggingface.co/datasets/allenai/ai2_arc}}

\begin{tcolorbox}[
  title=Example,
  colbacktitle=lightgray,
  coltitle=black
]
An astronomer observes that a planet rotates faster after a meteorite impact. Which is the most likely effect of this increase in rotation?\\
A. Planetary density will decrease.\\
B. Planetary years will become longer.\\
C. Planetary days will become shorter.\\
D. Planetary gravity will become stronger.\\
\end{tcolorbox}

\begin{table}[!h]
\centering
\small
\begin{tabular}{p{0.25\textwidth} p{0.69\textwidth}}
\toprule
    Answer format & Extraction logic \\
    \midrule
    \texttt{The answer is (<A/B/C/D/E>)} & The response is first uppercased. Extraction then tries, in order: the first match of \texttt{ANSWER IS (<A/B/C/D/E>)}, otherwise the first match of \texttt{ANSWER IS <A/B/C/D/E>}, otherwise the first standalone letter in \texttt{A-E} anywhere in the response. The reward is zero if no valid choice letter is found or if it does not equal the gold answer. \\
    \bottomrule
\end{tabular}
\end{table}

\paragraph{MMLU-Pro}
measures broad academic knowledge and reasoning across a wide range of professional and academic subjects.

Published under Apache License 2.0 in the official codebase of \citet{Wang:24}.

\begin{tcolorbox}[
  title=Example,
  colbacktitle=lightgray,
  coltitle=black
]
Typical advertising regulatory bodies suggest, for example, that adverts must not: encourage \_\_\_\_\_, cause unnecessary \_\_\_\_\_ or \_\_\_\_\_, and must not cause \_\_\_\_\_ offence.\\
A. Safe practices, Fear, Jealousy, Trivial\\
B. Unsafe practices, Distress, Joy, Trivial\\
C. Safe practices, Wants, Jealousy, Trivial\\
D. Safe practices, Distress, Fear, Trivial\\
E. Unsafe practices, Wants, Jealousy, Serious\\
F. Safe practices, Distress, Jealousy, Serious\\
G. Safe practices, Wants, Fear, Serious\\
H. Unsafe practices, Wants, Fear, Trivial\\
I. Unsafe practices, Distress, Fear, Serious\\
\end{tcolorbox}

\begin{table}[!h]
\centering
\small
\begin{tabular}{p{0.25\textwidth} p{0.69\textwidth}}
\toprule
    Answer format & Extraction logic \\
    \midrule
    \texttt{The answer is (<A/.../J>)}& The response is first uppercased. Extraction then tries, in order: the first match of \texttt{ANSWER IS (<A/.../J>)}, otherwise the first match of \texttt{ANSWER IS <A/.../J>}, otherwise the first standalone letter in \texttt{A-J} anywhere in the response. The reward is zero if no valid choice letter is found or if it does not equal the gold answer. \\
    \bottomrule
\end{tabular}
\end{table}

\FloatBarrier
\section{Detailed Results on Individual Prior-Task Accuracies} \label{app:detailed_results}

Accuracies for the base models on each task are provided in Tab.~\ref{tab:accuracies}.

\begin{table}[]
    \centering
    \small
    \setlength{\tabcolsep}{4pt}
    \caption{Individual task accuracies of original models before applying ES or GRPO. ES and GRPO implementations use different precisions in their original implementations, which was kept as is to stay consistent with prior work. ES: \texttt{float16}, GRPO: \texttt{bfloat16}. This leads to slight performance differences even for the base model under greedy decoding. Top: \texttt{float16} (ES). Bottom: \texttt{bfloat16} (GRPO). \vspace{0.1cm}}
    \label{tab:accuracies}
    \begin{tabular}{c|ccccccc}
    \toprule
    Model & Countdown & GSM8K & ProofWriter & HellaSwag & PIQA & ARC-C & MMLU-Pro \\
    \midrule
    Qwen-2.5 1.5B Instruct & $5.3$ & $58.2$ & $53.6$ & $57.5$ & $71.7$ & $75.3$ & $31.5$ \\
    Qwen-2.5 3B Instruct & $11.5$ & $83.8$ & $44.2$ & $63.7$ & $81.2$ & $85.2$ & $41.6$ \\
    Qwen-2.5 7B Instruct & $25.5$ & $91.2$ & $67.6$ & $83.1$ & $86.0$ & $91.1$ & $57.3$ \\
    Llama-3.2 3B Instruct & $4.0$ & $75.0$ & $38.4$ & $41.8$ & $66.4$ & $77.5$ & $20.4$ \\ 
    \midrule
    Qwen-2.5 1.5B Instruct & $4.9$ & $58.2$ & $51.9$ & $56.8$ & $71.4$ & $75.4$ & $32.1$  \\
    Qwen-2.5 3B Instruct & $10.0$ & $84.0$ & $45.1$ & $64.0$ & $81.3$ & $84.8$ & $41.9$ \\
    Qwen-2.5 7B Instruct & $ 24.5$ & $90.2$ & $67.9$ & $83.2$ & $86.1$ & $91.4$ & $56.9$ \\
    Llama-3.2 3B Instruct & $2.2$ & $75.2$ & $41.7$ & $40.6$ & $68.2$ & $78.6$ & $21.7$ \\
    \bottomrule
\end{tabular}
\end{table}

\begin{figure}[t]
    \centering
    \includegraphics[width=\linewidth]{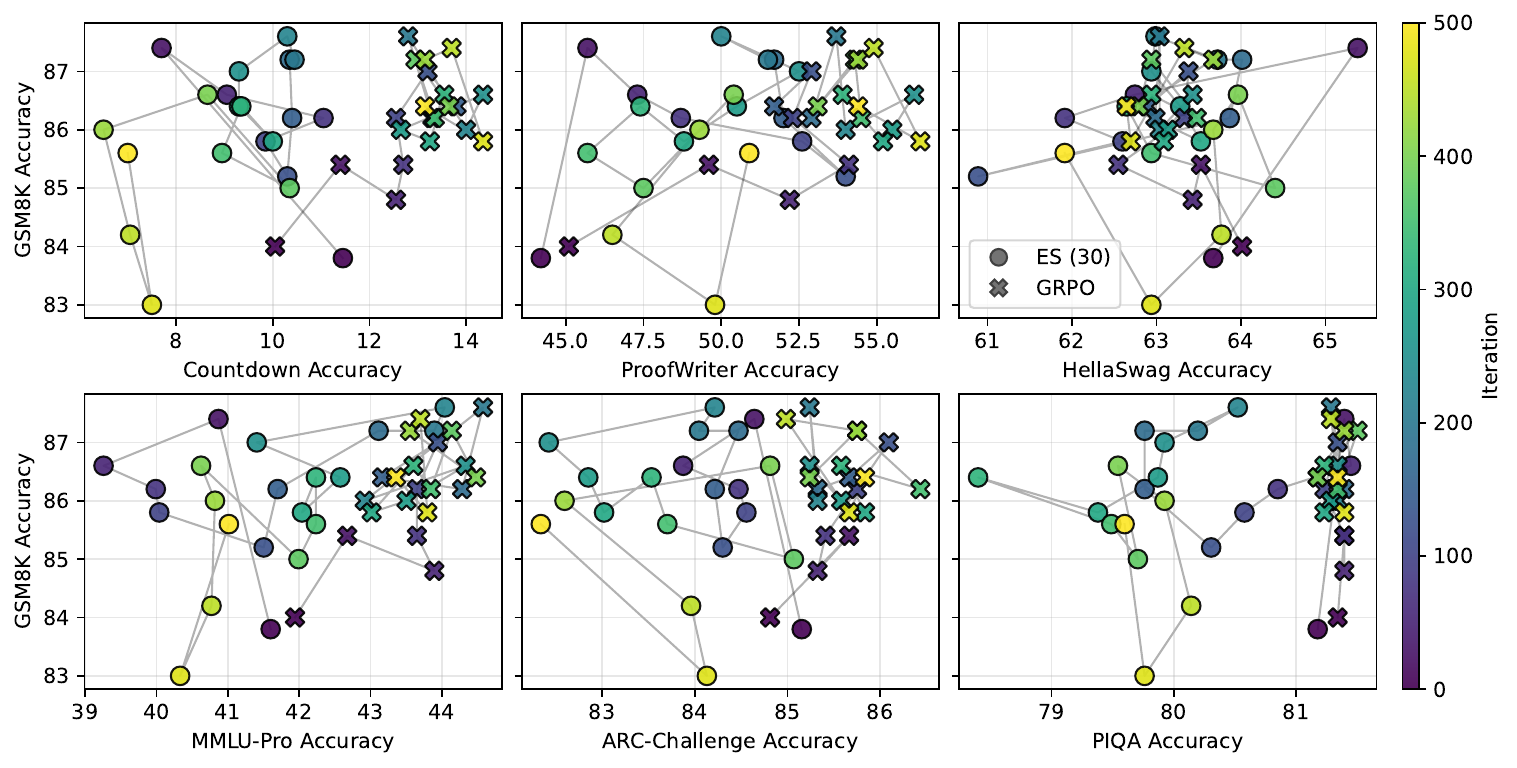}
    \caption{\emph{Individual prior task accuracies training on GSM8K as target task.} There is no systematic forgetting observed for ES nor for GRPO.}
    \label{fig:individual_gsm8k}
\end{figure}

\begin{figure}[t]
    \centering
    \includegraphics[width=\linewidth]{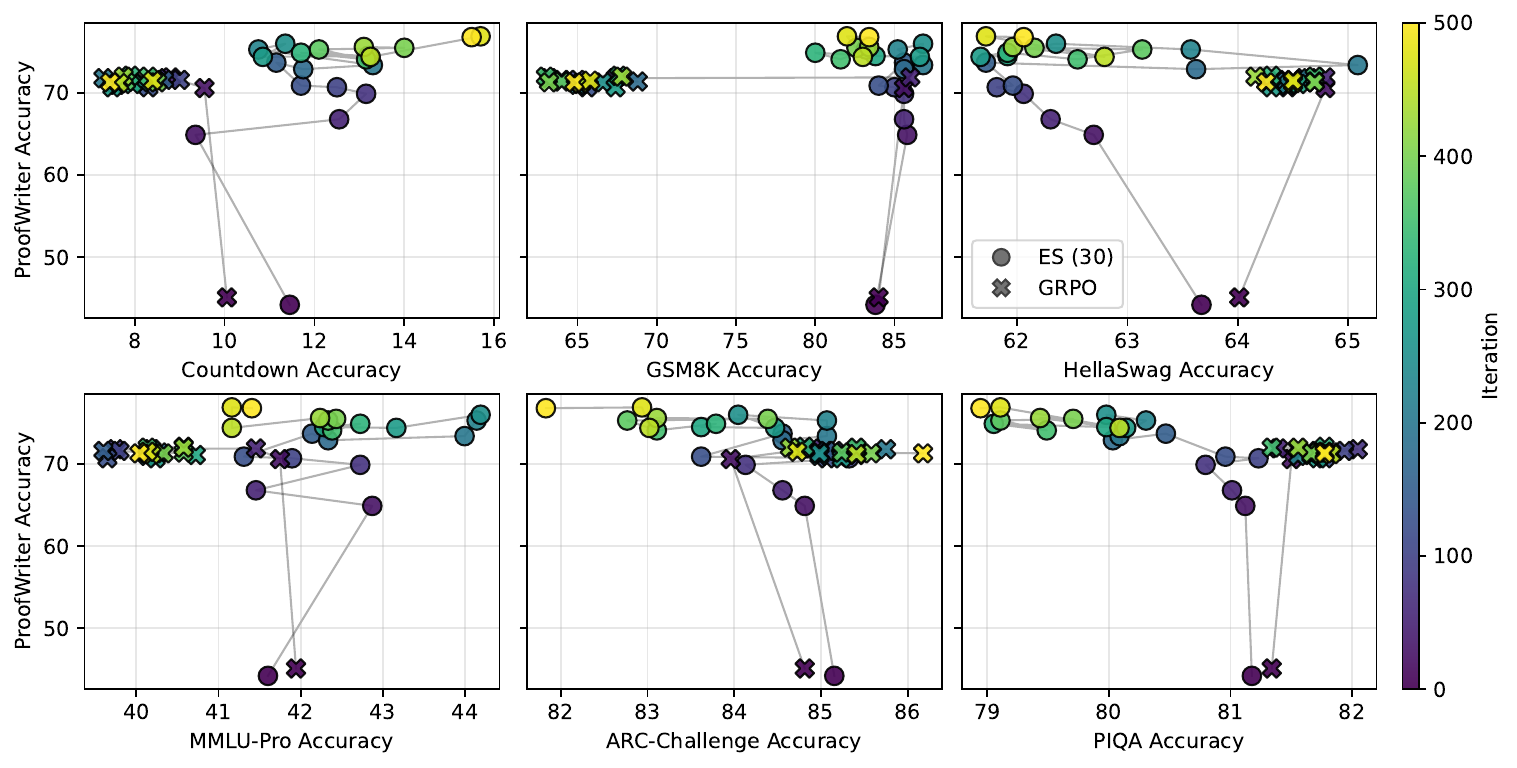}
    \caption{\emph{Individual prior task accuracies training on ProofWriter as target task.} Strong forgetting is observed for GRPO on GSM8K as a prior task, as well as moderate forgetting on Countdown and MMLU-Pro. For ES, mild forgetting is observed for HellaSwag, ARC-Challenge, and PIQA.}
    \label{fig:individual_proofwriter}
\end{figure}

\begin{figure}[t]
    \centering
    \includegraphics[width=\linewidth]{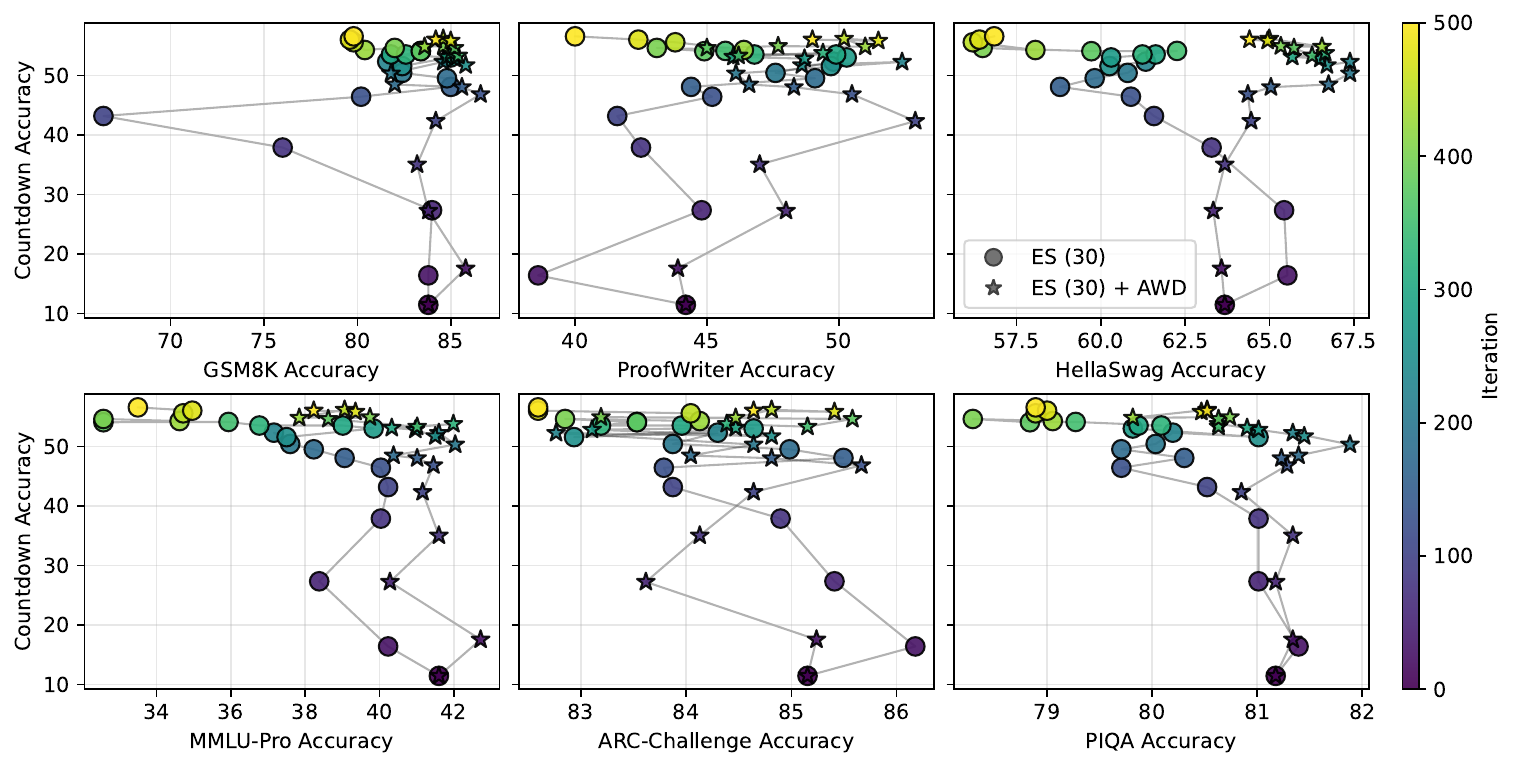}
    \caption{\emph{Individual prior task accuracies training on Countdown as target task with and without AWD.} While forgetting is observed for standard ES, there is essentially no forgetting in any task with ES + AWD ($\ell_2$), except a slight degradation on MMLU-Pro.}
    \label{fig:individual_awd}
\end{figure}

Furthermore, per-task accuracies for ES and GRPO optimization for GSM8K and ProofWriter as target tasks are shown in Fig.~\ref{fig:individual_gsm8k} and Fig.~\ref{fig:individual_proofwriter}.
Additionally, the individual performances per dataset of ES and ES+AWD, which was shown aggregated over datasets in Fig.~\ref{fig:main}, is shown in Fig.~\ref{fig:individual_awd}.

\section{Additional Results on Applying AWD} \label{app:detailed_awd}

This section provides additional results for applying AWD on different target datasets and model types as follows. 
Results for training on ProofWriter as target task with Qwen-2.5 3B Instruct as model are provided in Tab.~\ref{tab:awd_proofwriter}.
Results for training on Countdown as target task with Llama-3.2 3B Instruct as model are provided in Tab.~\ref{tab:awd_llama_countdown}.
Results for training on ProofWriter as target task with Llama-3.2 3B Instruct as model are provided in Tab.~\ref{tab:awd_llama_proofwriter}.

The penalty factor $\lambda$ was not tuned for individual configurations, showing that the values suggested by the ablation in the main paper---for $\ell_1$: $\lambda = 0.01$ and for $\ell_2$: $\lambda=10.0$---transfer well between different target task and model type configurations.

\begin{table}[t!]
    \centering
    \footnotesize
    \setlength{\tabcolsep}{3.3pt}
    \caption{\emph{Change in target (ProofWriter) and prior task accuracies using AWD with $\ell_1/\ell_2$ decay on Qwen-2.5 3B Instruct.} The average is calculated over accuracies in all prior tasks, thus excluding ProofWriter. Statistics are computed over three runs; for baseline accuracies see Tab.~\ref{tab:accuracies} in App~\ref{app:detailed_results}. For ProofWriter, the baseline without AWD already has relatively low forgetting. 
    Applying AWD with $\ell_1$ ($\lambda = 0.01$) leads to similar results, improving for some prior tasks, but leading to a bit more performance degradation for others. 
    Applying AWD with $\ell_2$ ($\lambda = 10.0$) leads to improvements for all prior tasks except Countdown (which has a high std). \vspace{0.1cm}}
    \label{tab:awd_proofwriter}
    \begin{tabular}{c|ccccccc|c}
    \toprule
    Run & Countdown & GSM8K & \hl{ProofWriter} & HellaSwag & PIQA & ARC-C & MMLU-Pro & Average \\
    \midrule
    ES (30) & $-0.6_{(4.8)}$ & $-0.3_{(0.3)}$ & $+35.6_{(2.7)}$ & $-1.4_{(0.2)}$ & $-1.9_{(0.8)}$ & $-2.5_{(1.6)}$ & $-0.7_{(0.5)}$ & $-1.3_{(0.8)}$ \\
    ES+AWD (L1) & $+0.4_{(3.0)}$ & $+0.7_{(0.8)}$ & $+34.6_{(0.8)}$ & $-3.6_{(2.2)}$ & $-1.1_{(0.5)}$ & $-1.0_{(0.3)}$ & $-2.7_{(3.5)}$ & $-1.2_{(0.4)}$ \\
    ES+AWD (L2) & $-2.3_{(3.4)}$ & $+1.3_{(0.6)}$ & $+34.4_{(0.4)}$ & $-1.1_{(3.1)}$ & $-0.9_{(1.1)}$ & $-0.5_{(0.2)}$ & $+0.5_{(0.8)}$ & $-0.5_{(0.6)}$ \\
    \bottomrule
    \end{tabular}
\end{table}

\begin{table}[t!]
    \centering
    \footnotesize
    \setlength{\tabcolsep}{3.3pt}
    \caption{\emph{Change in target (Countdown) and prior task accuracies using AWD with $\ell_1/\ell_2$ decay on Llama-3.2 3B Instruct .} The average is calculated over accuracies in all prior tasks, thus excluding Countdown. Statistics are computed over three runs; for baseline accuracies see Tab.~\ref{tab:accuracies} in App~\ref{app:detailed_results}. 
    Applying AWD with $\ell_1$ ($\lambda = 0.01$) leads to improvements across all prior tasks except ProofWriter, which has relatively high variance.
    Applying AWD with $\ell_2$ ($\lambda = 10.0$) leads to even stronger improvements generally, again except for ProofWriter where it decreases prior task performance.\vspace{0.1cm}}
    \label{tab:awd_llama_countdown}
    \begin{tabular}{c|ccccccc|c}
    \toprule
    Run & \hl{Countdown} & GSM8K & ProofWriter & HellaSwag & PIQA & ARC-C & MMLU-Pro & Average \\
    \midrule
    ES (30) & $+46.9_{(1.0)}$ & $-5.1_{(3.4)}$ & $+3.9_{(6.0)}$ & $-12.6_{(2.0)}$ & $-3.3_{(3.3)}$ & $-4.2_{(1.1)}$ & $-5.4_{(3.8)}$ & $-4.4_{(1.7)}$ \\
    ES+AWD (L1) & $+45.0_{(0.4)}$ & $+1.9_{(2.1)}$ & $-0.1_{(5.1)}$ & $-4.0_{(0.9)}$ & $-1.6_{(2.2)}$ & $-1.7_{(0.3)}$ & $-5.3_{(2.5)}$ & $-1.8_{(1.0)}$ \\
    ES+AWD (L2) & $+44.6_{(0.1)}$ & $+2.9_{(2.1)}$ & $-1.7_{(3.6)}$ & $-2.8_{(0.7)}$ & $+1.6_{(2.2)}$ & $-0.9_{(0.6)}$ & $-3.5_{(3.5)}$ & $-0.7_{(1.0)}$ \\
    \bottomrule
    \end{tabular}
\end{table}

\begin{table}[t!]
    \centering
    \footnotesize
    \setlength{\tabcolsep}{3.3pt}
    \caption{\emph{Change in target (ProofWriter) and prior task accuracies using AWD with $\ell_1/\ell_2$ decay for Llama-3.2 3B Instruct.} The average is calculated over accuracies in all prior tasks, thus excluding ProofWriter. Statistics are computed over three runs; for baseline accuracies see Tab.~\ref{tab:accuracies} in App~\ref{app:detailed_results}.
    Applying AWD with $\ell_1$ ($\lambda = 0.01$) mostly decreases prior task forgetting, especially on HellaSwag, ARC and MMLU-Pro where it was most severe without AWD.
    Applying AWD with $\ell_2$ ($\lambda = 10.0$) leads to similar improvements, yet also a substantial drop for PIQA - with high variance. \vspace{0.1cm}}
    \label{tab:awd_llama_proofwriter}
    \begin{tabular}{c|ccccccc|c}
    \toprule
    Run & Countdown & GSM8K & \hl{ProofWriter} & HellaSwag & PIQA & ARC-C & MMLU-Pro & Average \\
    \midrule
    ES (30) & $+0.8_{(1.7)}$ & $-4.7_{(0.8)}$ & $+30.5_{(1.6)}$ & $-5.4_{(4.5)}$ & $-3.0_{(2.5)}$ & $-6.7_{(0.6)}$ & $-4.4_{(7.2)}$ & $-3.9_{(2.4)}$ \\
    ES+AWD (L1) & $+0.4_{(0.8)}$ & $-1.1_{(2.7)}$ & $+30.8_{(1.6)}$ & $-0.1_{(2.4)}$ & $-3.4_{(4.8)}$ & $-1.9_{(0.8)}$ & $-3.7_{(3.3)}$ & $-1.6_{(1.7)}$ \\
    ES+AWD (L2) & $+0.9_{(0.7)}$ & $-0.6_{(1.3)}$ & $+29.3_{(1.4)}$ & $-2.1_{(4.0)}$ & $-4.5_{(4.5)}$ & $-1.0_{(1.0)}$ & $+2.9_{(4.2)}$ & $-0.8_{(1.7)}$ \\
    \bottomrule
    \end{tabular}
\end{table}

\section{KL-Divergence for Different Training Tasks}

\begin{figure}[]
    \centering
    \includegraphics[width=\linewidth]{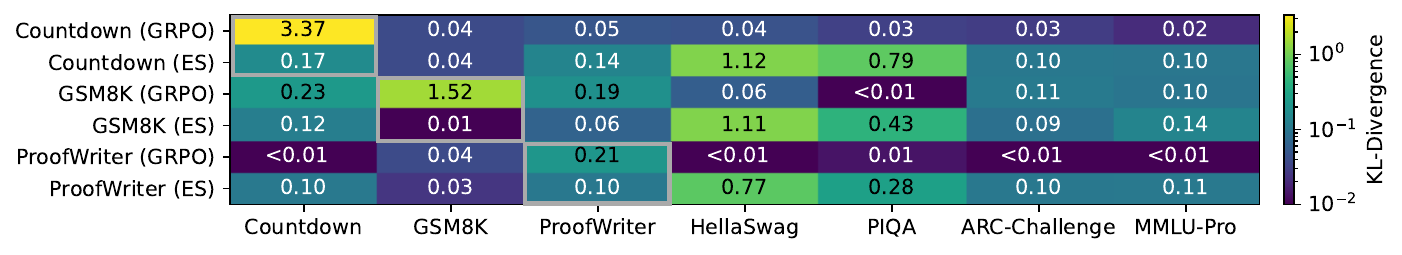}
    \caption{\emph{KL-divergence $\text{KL}(p_{\btheta_0} || p_{\btheta_T})$ for different target tasks across prior tasks for ES (population size 30) and GRPO.} The y-axis shows training task and fine-tuning method, the x-axis shows the task used for evaluation. When evaluating on the target task (grey boxes), ES (30) attains lower KL-divergence than GRPO, though both have similar accuracies. For the prior-tasks, GRPO mostly has lower KL-Divergences then ES (30).}
    \label{fig:kl_tasks}
\end{figure}

In Sec.~\ref{sec:analysis} in the main paper, the KL-divergence on the target task Countdown and the remaining prior tasks was evaluated for ES under different population sizes, as well as when combined with AWD.
Complementary to those results, Fig.~\ref{fig:kl_tasks} shows the KL-divergence for GRPO and ES with population size 30 for the three different target tasks.
In line with \citet{Qiu:25}, the results show lower KL-divergence for ES when evaluated on the target task.
For the prior-tasks not trained on, GRPO generally attains lower KL-divergence than ES with low population size.
This might be explained by the more targeted, low-magnitude changes induced by GRPO.
For the target task, those lead to massive changes in the output distribution, while they less impact the output distribution of dissimilar tasks, while ES changes the output distribution less for the particular tasks, but induces more global change.
Notably though, the results in the main paper showed that the KL-divergence of ES on prior-tasks reduces with higher population size and when combined with AWD.

\section{Detailed Results on Dependence on Target Task and Model Type} \label{app:task_model_type}

Detailed results comparing ES with a population size of 30 with GRPO are provided as follows.
Tab.~\ref{tab:model_families_cd} shows accuracy change across model families (for the 3B size each), when trained on Countdown as target task.
Tab.~\ref{tab:model_families_pw} shows accuracy change across model families (for the 3B size each), when trained on ProofWriter as target task.
Tab.~\ref{tab:model_size_cd} shows accuracy change across model size (for Qwen model series), when trained on Countdown as target task.
Tab.~\ref{tab:model_size_pw} shows accuracy change across model size (for Qwen model series), when trained on ProofWriter as target task.

\section{Compute Resources} \label{app:compute}

All experiments were performed on a node with eight H200s.
Individual runs for ES on the 3B model on Countdown took around 4 node-hours, the smaller 1.5B model took around three node-hours, and the larger 7B model around six node-hours.
For GRPO, a 3B run took around 20 node-hours, a 1.5B run around 14 node-hours, and a 7B run took around 32 node-hours.
For ES, runs for GSM8K and ProofWriter require only around 75\% of the training time of Countdown, as answer lengths are generally shorter (generating answers for the population is the dominant compute cost for ES).
For GRPO, this discrepancy is much smaller as generating the answers is only one part of the overall compute cost, with forward and backward pass for updating the model as well as another forward pass for the KL-regularization being more dominant, leading to more uniform runtimes over different tasks.

Overall, the tracked node-hours for the whole project, including failed runs and initial experiments not included in the final paper, amount to approximately 52 node-days (1248 node-hours).
Performing evaluations on the trained models took approximately one extra node-day.

\clearpage

\begin{table}[]
\centering
\caption{\emph{Accuracy change across model families.} The target task is Countdown, and the average is calculated over all other tasks. Statistics are computed over three runs; for baseline accuracies see Tab.~\ref{tab:accuracies} in App~\ref{app:detailed_results}. Prior-task forgetting is similar for both the Qwen and Llama 3B model. \vspace{0.1cm}}
\label{tab:model_families_cd}
\footnotesize
\setlength{\tabcolsep}{2.4pt}
\begin{tabular}{c|ccccccc|c}
\toprule
Method & \hl{Countdown} & GSM8K & ProofWriter & HellaSwag & PIQA & ARC-C & MMLU-Pro & Average \\
\midrule
Qwen2.5 (ES) & $+45.3_{(0.3)}$ & $-9.5_{(11.5)}$ & $-2.0_{(5.7)}$ & $-8.2_{(9.0)}$ & $-2.8_{(1.0)}$ & $-2.0_{(0.5)}$ & $-6.5_{(2.6)}$ & $-5.2_{(4.7)}$ \\
Qwen2.5 (GRPO) & $+46.9_{(2.4)}$ & $+0.5_{(0.9)}$ & $+3.5_{(4.3)}$ & $+1.1_{(0.2)}$ & $+0.2_{(0.2)}$ & $+0.1_{(0.4)}$ & $-2.3_{(1.7)}$ & $+0.5_{(0.9)}$ \\ \midrule
Llama-3.2 (ES) & $+46.9_{(1.0)}$ & $-5.1_{(3.4)}$ & $+3.9_{(6.0)}$ & $-12.6_{(2.0)}$ & $-3.3_{(3.3)}$ & $-4.2_{(1.1)}$ & $-5.4_{(3.8)}$ & $-4.4_{(1.7)}$ \\
Llama-3.2 (GRPO) & $+41.5_{(9.4)}$ & $-0.4_{(1.6)}$ & $+1.2_{(0.3)}$ & $+1.3_{(1.3)}$ & $-7.6_{(4.5)}$ & $+0.5_{(0.4)}$ & $-2.8_{(2.8)}$ & $-1.3_{(1.0)}$ \\
\bottomrule
\end{tabular}
\end{table}

\begin{table}[]
\centering
\caption{\emph{Accuracy change across model families.} The target task is ProofWriter, and the average is calculated over all other tasks. Statistics are computed over three runs; for baseline accuracies see Tab.~\ref{tab:accuracies} in App~\ref{app:detailed_results}. Prior-task forgetting is similar for ES, but differs for GRPO. \vspace{0.1cm}}
\label{tab:model_families_pw}
\footnotesize
\setlength{\tabcolsep}{2.4pt}
\begin{tabular}{c|ccccccc|c}
\toprule
Method & Countdown & GSM8K & \hl{ProofWriter} & HellaSwag & PIQA & ARC-C & MMLU-Pro & Average \\
\midrule
Qwen2.5 (ES) & $-0.6_{(4.8)}$ & $-0.3_{(0.3)}$ & $+35.6_{(2.7)}$ & $-1.4_{(0.2)}$ & $-1.9_{(0.8)}$ & $-2.5_{(1.6)}$ & $-0.7_{(0.5)}$ & $-1.3_{(0.8)}$ \\
Qwen2.5 (GRPO) & $-2.3_{(0.6)}$ & $-10.1_{(9.0)}$ & $+25.3_{(1.5)}$ & $-0.8_{(0.9)}$ & $-0.0_{(0.4)}$ & $+0.6_{(0.8)}$ & $-1.0_{(0.8)}$ & $-2.3_{(1.4)}$ \\ \midrule
Llama-3.2 (ES) & $+0.8_{(1.7)}$ & $-4.7_{(0.8)}$ & $+30.5_{(1.6)}$ & $-5.4_{(4.5)}$ & $-3.0_{(2.5)}$ & $-6.7_{(0.6)}$ & $-4.4_{(7.2)}$ & $-3.9_{(2.4)}$ \\
Llama-3.2 (GRPO) & $+3.8_{(2.2)}$ & $-0.3_{(2.6)}$ & $+24.2_{(1.2)}$ & $+6.4_{(2.7)}$ & $+0.4_{(1.5)}$ & $-0.1_{(0.2)}$ & $+1.3_{(6.2)}$ & $+1.9_{(0.9)}$ \\
\bottomrule
\end{tabular}
\end{table}

\begin{table}
\centering
\caption{\emph{Accuracy change across different model sizes.} The target task is Countdown and the average is calculated over all other tasks. Statistics are computed over three runs; for baseline accuracies see Tab.~\ref{tab:accuracies} in App~\ref{app:detailed_results}. Forgetting of ES arises under all model sizes, whereas GRPO does not exhibit pronounced forgetting. \vspace{0.1cm}}
\footnotesize
\setlength{\tabcolsep}{3.8pt}
\label{tab:model_size_cd}
\begin{tabular}{c|ccccccc|c}
\toprule
Method & \hl{Countdown} & GSM8K & ProofWriter & HellaSwag & PIQA & ARC-C & MMLU-Pro & Average \\
\midrule
ES 1.5B & $+36.9_{(9.8)}$ & $+4.5_{(5.0)}$ & $-3.9_{(2.8)}$ & $-9.7_{(3.4)}$ & $-3.2_{(1.9)}$ & $-4.2_{(5.1)}$ & $-2.8_{(1.1)}$ & $-3.2_{(0.8)}$ \\
ES 3B & $+45.3_{(0.3)}$ & $-9.5_{(11.5)}$ & $-2.0_{(5.7)}$ & $-8.2_{(9.0)}$ & $-2.8_{(1.0)}$ & $-2.0_{(0.5)}$ & $-6.5_{(2.6)}$ & $-5.2_{(4.7)}$ \\
ES 7B & $+35.0_{(2.5)}$ & $-0.8_{(1.2)}$ & $-8.2_{(8.1)}$ & $-3.3_{(1.2)}$ & $-2.0_{(0.5)}$ & $-1.1_{(0.4)}$ & $-4.8_{(1.3)}$ & $-3.4_{(1.6)}$ \\
GRPO 1.5B & $+17.3_{(0.1)}$ & $+16.8_{(1.0)}$ & $+1.0_{(0.5)}$ & $-1.1_{(1.2)}$ & $+1.9_{(1.0)}$ & $-0.3_{(0.1)}$ & $+0.4_{(0.4)}$ & $+3.1_{(0.2)}$ \\
GRPO 3B & $+46.9_{(2.4)}$ & $+0.5_{(0.9)}$ & $+3.5_{(4.3)}$ & $+1.1_{(0.2)}$ & $+0.2_{(0.2)}$ & $+0.1_{(0.4)}$ & $-2.3_{(1.7)}$ & $+0.5_{(0.9)}$ \\
GRPO 7B & $+36.8_{(0.3)}$ & $+2.5_{(0.9)}$ & $-0.3_{(0.3)}$ & $+0.8_{(0.5)}$ & $+0.3_{(0.2)}$ & $+0.2_{(0.3)}$ & $-0.5_{(0.1)}$ & $+0.5_{(0.2)}$ \\
\bottomrule
\end{tabular}
\end{table}

\begin{table}
\centering
\caption{\emph{Accuracy change across different model sizes.} The target task is ProofWriter and the average is calculated over all other tasks. Statistics are computed over three runs; for baseline accuracies see Tab.~\ref{tab:accuracies} in App~\ref{app:detailed_results}. Forgetting of ES arises under all model sizes. GRPO also exhibits forgetting for the two larger model sizes. \vspace{0.1cm}}
\footnotesize
\setlength{\tabcolsep}{3.8pt}
\label{tab:model_size_pw}
\begin{tabular}{c|ccccccc|c}
\toprule
Method & Countdown & GSM8K & \hl{ProofWriter} & HellaSwag & PIQA & ARC-C & MMLU-Pro & Average \\
\midrule
ES 1.5B & $-0.9_{(2.6)}$ & $+0.7_{(16.1)}$ & $+9.8_{(0.4)}$ & $-7.1_{(2.7)}$ & $-0.9_{(2.8)}$ & $-2.0_{(0.8)}$ & $-2.7_{(1.7)}$ & $-2.2_{(2.5)}$ \\
ES 3B & $-0.6_{(4.8)}$ & $-0.3_{(0.3)}$ & $+35.6_{(2.7)}$ & $-1.4_{(0.2)}$ & $-1.9_{(0.8)}$ & $-2.5_{(1.6)}$ & $-0.7_{(0.5)}$ & $-1.3_{(0.8)}$ \\
ES 7B & $-2.5_{(3.0)}$ & $-1.0_{(0.7)}$ & $+10.9_{(1.7)}$ & $-3.5_{(1.2)}$ & $-2.2_{(0.6)}$ & $-0.7_{(0.8)}$ & $-5.0_{(0.8)}$ & $-2.5_{(0.8)}$ \\
GRPO 1.5B & $-0.9_{(2.7)}$ & $+7.8_{(4.2)}$ & $+12.4_{(3.5)}$ & $-0.2_{(1.3)}$ & $+0.9_{(0.4)}$ & $-0.1_{(0.3)}$ & $+0.3_{(0.9)}$ & $+1.3_{(0.3)}$ \\
GRPO 3B & $-2.3_{(0.6)}$ & $-10.1_{(9.0)}$ & $+25.3_{(1.5)}$ & $-0.8_{(0.9)}$ & $-0.0_{(0.4)}$ & $+0.6_{(0.8)}$ & $-1.0_{(0.8)}$ & $-2.3_{(1.4)}$ \\
GRPO 7B & $+0.8_{(1.2)}$ & $+1.5_{(1.2)}$ & $+7.2_{(6.7)}$ & $-0.6_{(1.8)}$ & $-1.3_{(1.6)}$ & $+0.0_{(0.4)}$ & $-0.9_{(0.9)}$ & $-0.1_{(1.0)}$ \\
\bottomrule
\end{tabular}
\end{table}

\end{document}